\documentclass[10pt,journal,compsoc]{IEEEtran}

\usepackage{amsmath,amssymb,mathtools,bm}
\usepackage{booktabs,multirow,array}
\usepackage{graphicx}
\usepackage{dblfloatfix}
\usepackage{placeins}
\usepackage{xcolor}
\usepackage{cite}
\usepackage{microtype}
\usepackage{url}
\usepackage{tikz}
\usetikzlibrary{positioning,arrows.meta,calc,fit}

\newcommand{\Istruct}{I_{\mathrm{struct}}}

\title{A Geometry-Based Capacity Theory for Finite-Feature Associative Memory}

\author{Jianhai~Zhang\textsuperscript{*}, Donghao~Zhang, Pattarawut~Charatpangoon, Bijoy~Menon, M.~Ethan~MacDonald, Wu~Qiu, and Aravind~Ganesh\textsuperscript{*}%
\IEEEcompsocitemizethanks{%
\IEEEcompsocthanksitem J. Zhang, P. Charatpangoon, B. Menon, and A. Ganesh are with the Department of Clinical Neurosciences, Cumming School of Medicine, University of Calgary, and the Ron and Rene Ward Centre for Healthy Brain Aging Research, Hotchkiss Brain Institute, University of Calgary. \textsuperscript{*}Corresponding authors: Jianhai Zhang and Aravind Ganesh.
\IEEEcompsocthanksitem D. Zhang and W. Qiu are with the Department of Biomedical Engineering, School of Life Science and Technology, Huazhong University of Science and Technology, and the Advanced Biomedical Imaging Facility, Huazhong University of Science and Technology.
\IEEEcompsocthanksitem M. E. MacDonald is with the Department of Biomedical Engineering, Schulich School of Engineering, University of Calgary, and the Ron and Rene Ward Centre for Healthy Brain Aging Research, Hotchkiss Brain Institute, University of Calgary.}}
\begin{document}
\maketitle

\begin{abstract}
We develop a geometry-based capacity theory for exact-key retrieval in compressed finite-feature Hebbian associative memory. For random or approximately isotropic values, retrieval interference separates into finite-feature noise, which decreases with feature dimension, and structural interference, which is determined by squared kernel overlap among stored keys and persists in the infinite-feature limit. This yields a fit-free prediction of retrieval quality, reveals a geometry-dependent capacity ceiling, and predicts the feature budget required for a target retrieval quality. When stored values are correlated, we show that retrieval depends jointly on the key kernel and value Gram matrix, and derive finite-feature approximations that account for this interaction. We validate the theory on synthetic, visual, and medical-image representations. Overall, the framework links representation geometry directly to memory capacity and distinguishes when performance can be improved by increasing the feature budget and when the representation itself must be changed. Across these settings, the predicted retrieval curves closely match empirical behavior and correctly identify changes in the preferred memory design.
\end{abstract}

\begin{IEEEkeywords}
Associative memory, Hebbian memory, random features, representation geometry, kernel methods, memory capacity, visual representations, medical imaging.
\end{IEEEkeywords}

\section{Introduction}\label{sec:introduction}

\IEEEPARstart{A}{ssociative} memory retrieves stored information from a content-based query rather than from a fixed address. Classical distributed and recurrent memories characterize storage limits through interference among stored patterns \cite{willshaw1969,little1974,hebb1949,hopfield1982,mceliece1987,amit1985spin,amit1985storage,tsodyks1988}. Dense and modern Hopfield models change the interaction or separation function and therefore operate in different storage regimes, while attention-based variants retain explicit key--value items \cite{krotov2016dense,demircigil2017huge,krotov2021large,ramsauer2021hopfield,millidge2022universal}. In this work, capacity refers to how much distinct feature information can be represented and retained before interference substantially degrades retrieval or discrimination. Rather than treating capacity as an abstract storage quantity, we define it operationally through how many separable feature patterns can be preserved while maintaining acceptable retrieval or classification quality. In practical AI systems, this asks whether increasingly diverse image, patient, or other sample representations can be compressed into a fixed memory without losing the information that distinguishes them.

This paper focuses on a compressed finite-feature regime. A collection of key--value pairs is aggregated into a fixed Hebbian matrix through additive outer products of nonlinear random-feature addresses. This storage rule is closely related to classical correlation-matrix memories and to fast-weight or linear-attention memories \cite{anderson1972interactive,kohonen1972correlation,schmidhuber1992fastweights,ba2016fastweights,katharopoulos2020linear,schlag2021fastweights}. Prior work has already shown that finite-dimensional fast-weight memories degrade as many nonorthogonal keys are accumulated. Recent and concurrent work studies related aspects of this problem through self-recall diagnostics and sparse caching in linear attention, spherical-packing and Welch-bound analyses of linear-attention capacity, sharp pattern-count thresholds for linear associative memories, and Hebbian fact-storage models that allow arbitrary input/output geometry \cite{mcdermott2026lola,ghriss2026kata,barnfield2026sharp,giorlandino2026sharp,garcia2026hebbians}. We therefore do not claim that nonorthogonality, geometry dependence, or capacity limits in linear memories are new phenomena. Our focus is narrower: we derive a quantitative item-wise law that uses the measured kernel geometry of the stored representations, separates deterministic overlap from finite-feature approximation noise, and remains applicable when stored values are correlated. We study exact-key retrieval, $q=k_i$, so that the analysis isolates cross-talk caused by compressed storage. Throughout, \emph{capacity} is operational: for a fixed set of $N$ stored keys, it refers to the feature budget $R/N$ required to reach a specified retrieval cosine, or equivalently the retrieval cosine attainable at a fixed budget. In practical terms, this asks: for a given representation and target retrieval quality, how large must the compressed memory be, and when does increasing the feature budget stop helping because the remaining interference is caused by the representation geometry itself? This is not an asymptotic pattern-count capacity claim for recurrent or explicit-storage memories. Robustness to corrupted, partial, or shifted queries is outside the scope of the present analysis.

Random Fourier features (RFFs) provide a convenient finite-dimensional approximation to shift-invariant kernels \cite{rahimi2007random}. Their approximation error has been studied extensively through structured constructions, variance reduction, and statistical guarantees \cite{le2013fastfood,yu2016orf,sutherland2015error,liu2022survey}. In a compressed associative memory, however, accurate kernel approximation does not remove all cross-talk. If stored keys remain similar under the kernel, off-target values still contribute to retrieval even in the infinite-feature limit. This motivates separating two sources of error: finite-feature noise and structural interference caused by the geometry of the stored representations. This distinction is practically useful because the two failure modes require different remedies. Finite-feature noise can be reduced by increasing the feature budget, whereas structural interference requires changing the representation, kernel, or storage organization.

This structural term is especially relevant for learned visual representations, whose pairwise geometry can vary substantially with architecture, training objective, centering, whitening, and kernel scale \cite{cogswell2016decov,ermolov2021whitening,zbontar2021barlow,bardes2022vicreg,jing2022collapse}. Effective rank summarizes spectral concentration \cite{roy2007effectiverank,garrido2023rankme}, but it does not describe the pairwise similarities seen by a particular memory kernel. We therefore work directly with the kernel matrix of the stored keys. For random or approximately isotropic values, the leading approximation separates retrieval interference into a finite-feature term, approximately $(N-1)/R$, and a structural term $\sum_{j\neq i}K_{ij}^{2}$. Intuitively, the structural term measures how strongly the other stored items overlap with item $i$ under the chosen kernel. Once the kernel, stored keys, number of items $N$, and feature budget $R$ are specified, no coefficient is fitted to the measured retrieval curve. This matters because two representations with the same nominal dimension, or even a similar effective rank, can require very different memory budgets if their pairwise kernel geometry differs.

We first test this law on i.i.d., clustered, low-rank, and anisotropic synthetic keys, and then on learned visual representations. ResNet50 and ViT-B/16 embeddings \cite{he2016resnet,dosovitskiy2021vit} are evaluated under RAW, CENTERED, and PCA-WHITENED geometry interventions and across multiple kernel scales. CIFAR-10 and STL-10 provide additional data and background conditions for fit-free boundary prediction, while DINOv2 ViT-B/14 \cite{oquab2024dinov2} and CLIP ViT-B/16 \cite{radford2021clip} extend the analysis to self-supervised and vision-language representations. As a public structured-domain stress test, we use the SynthRAD2023 Task-1 MRI--CT training data \cite{thummerer2023synthrad}, comprising 180 brain and 180 pelvis pairs. We then use 445 spatially aligned brain NCCT--CTA pairs as an independent clinical cohort. These experiments test whether the same analytic dependence on measured kernel geometry holds across substantially different representation regimes. The medical settings are especially informative because anatomically similar observations, adjacent slices, or patient-derived representations may coexist in the same finite memory. Their structured similarity can increase interference even when the nominal feature dimension is large.

Real learned values require an additional step. When values are correlated, the infinite-feature retrieval vector becomes a kernel-weighted mixture of stored values, so its target cosine depends jointly on the key kernel and the value Gram matrix. We characterize this deterministic joint key--value geometry exactly and approximate only the finite-$R$ perturbation. Aligned, permuted, and random-value controls then test whether value similarity is consistent with the neighborhoods defined by the key kernel. Repeated permutations are used to measure the stability of this effect rather than relying on a single random shuffle.

\begin{figure*}[!t]
\centering
\includegraphics[width=0.8\linewidth]{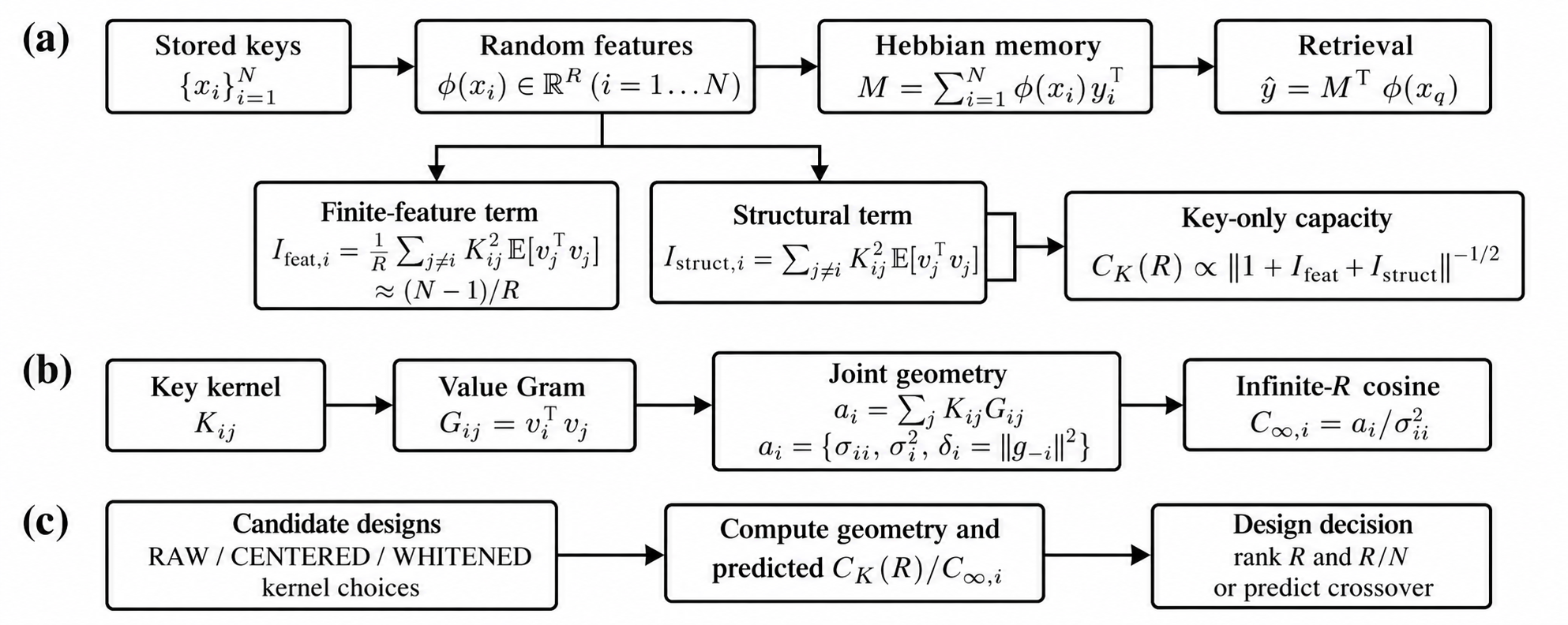}
\caption{Overview of the analysis. (a) In a finite random-feature Hebbian memory, retrieval error separates into a finite-feature term that can be reduced by increasing the memory budget and a structural-interference term imposed by representation geometry. (b) With correlated values, the relevant deterministic quantities are the kernel-weighted value mixture and the associated value Gram geometry. (c) The resulting predicted capacity curves can be compared across preprocessing and kernel choices to estimate the required memory size, rank designs, or predict a crossover.}
\label{fig:framework}
\end{figure*}

Finally, we ask whether the theory can guide memory design rather than only explain observed retrieval behavior. Candidate preprocessing and kernel choices are ranked using their predicted capacity curves, and the predicted RAW-to-WHITENED transition is compared with the empirical crossover. More broadly, the framework separates two different remedies for poor retrieval: increasing the finite feature budget when approximation noise dominates, or changing the representation when structural interference sets the limiting ceiling. A covariance-aware finite-$R$ term is not uniformly better for absolute-capacity prediction, but in our aligned real-value experiments it improves crossover prediction. The main contribution is therefore analytical rather than architectural: a geometry-dependent capacity law for compressed finite-feature memory, an extension to correlated values through joint key--value geometry, and a test of whether the same quantities can guide memory sizing and representation selection. Figure~\ref{fig:framework} summarizes this progression.

\section{Related Work}

\subsection{Classical associative and correlation-matrix memories}
Associative memory predates modern deep learning, with early distributed memories \cite{willshaw1969}, persistent-state models \cite{little1974}, and Hebbian learning \cite{hebb1949}. Particularly close to the storage rule studied here are early linear heteroassociative or correlation-matrix memories, which store key--value associations by additive outer products and retrieve a value by multiplying the resulting matrix by a cue \cite{anderson1972interactive,kohonen1972correlation}. Hopfield networks introduced recurrent attractor dynamics \cite{hopfield1982}, and classical statistical-mechanical and coding-theoretic analyses quantified storage transitions for random patterns \cite{amit1985spin,amit1985storage,mceliece1987}. Sparse coding can substantially alter capacity scaling \cite{tsodyks1988}, while sparse distributed memory provides a high-dimensional alternative \cite{kanerva1988}. Dense associative memories increase capacity by modifying the interaction or energy function \cite{krotov2016dense,demircigil2017huge}, with later work relating large-capacity models to hidden-unit constructions \cite{krotov2021large}. Our setting is closest to correlation-matrix memory, but replaces the original address vectors with a finite random-feature map and asks how the resulting kernel geometry controls cross-talk and exact-key retrieval quality.

\subsection{Fast-weight memories and linear attention}
Fast-weight models use rapidly changing matrices to store recent associations without retaining an explicit list of all past activations \cite{schmidhuber1992fastweights,ba2016fastweights}. This viewpoint is especially relevant to linear attention: kernelized linear Transformers rewrite attention as a feature-map outer-product memory that can be updated recurrently \cite{katharopoulos2020linear}, and Schlag \emph{et al.} showed the formal equivalence between such linearized attention and fast-weight programmers \cite{schlag2021fastweights}. Their analysis also identifies a capacity limitation when too many nonorthogonal key features are accumulated in a finite-dimensional fast-weight matrix. Our analysis refines this dimension/orthogonality picture into a graded prediction based on the measured overlap $\sum_{j\neq i}K_{ij}^2$, while separately accounting for finite-feature approximation noise. Performer \cite{choromanski2021performer} and Random Feature Attention \cite{peng2021rfa} further use random features to approximate attention kernels efficiently. Their primary goal is efficient approximation of normalized attention, whereas we study exact-key retrieval quality of a fixed compressed Hebbian matrix. More recent work makes the memory-collision issue still more explicit. LoLA uses self-recall error to identify key--value pairs that a recurrent linear-attention state remembers poorly and protects them with a sparse cache \cite{mcdermott2026lola}. KATA formulates associative recall as a spherical-packing problem and characterizes a capacity--interference trade-off through a Welch interference floor while designing new feature maps \cite{ghriss2026kata}. Complementary capacity analyses derive sharp pattern-count thresholds for linear associative memories under winner-take-all or strict-separation criteria \cite{barnfield2026sharp,giorlandino2026sharp}. Concurrent Hebbian constructions for Transformer MLPs analyze decoding-margin storage scaling under arbitrary key and value geometries \cite{garcia2026hebbians}. These results are closely related in mechanism or terminology, but address different observables and capacity notions. To our knowledge, they do not provide an item-wise, kernel-conditional normalized retrieval-cosine law for compressed finite-random-feature Hebbian memory that separates finite-feature approximation noise from deterministic key overlap, extends quantitatively to correlated values, and predicts finite-memory design choices directly from measured representation geometry. Our narrower question is therefore how measured key geometry determines exact-key retrieval cross-talk in this compressed storage rule, how finite-feature noise enters separately, and how the analysis changes for correlated values.

\subsection{Explicit neural memories and modern Hopfield networks}
Memory Networks and end-to-end variants introduced learned read/write mechanisms over explicit external memories \cite{weston2015memory,sukhbaatar2015end}; product-key memories provide efficient access to very large learned tables \cite{lample2019product}. Transformer attention \cite{vaswani2017attention} and modern continuous Hopfield networks are closely related mathematically \cite{ramsauer2021hopfield}, while Universal Hopfield Networks cast several memory models as similarity--separation--projection systems \cite{millidge2022universal}; sparse distributed memory has also been connected to attention \cite{bricken2021attention}. These methods typically retain, address, or parameterize individual memory items and therefore differ from the fixed-size compressed matrix analyzed here. We use them as contextual baselines rather than as storage-equivalent comparators.

\subsection{Random features and scalable kernel approximation}
Kernel methods provide a natural language for nonlinear similarity \cite{scholkopf2002learning}. Low-rank Nystr\"om approximations \cite{williams2001nystrom} and random Fourier features (RFFs) \cite{rahimi2007random} are canonical routes to scalable kernel learning. Subsequent work developed structured fast transforms \cite{le2013fastfood}, quasi-Monte Carlo maps \cite{yang2014qmc}, orthogonal random features \cite{yu2016orf}, sharper approximation-error analyses \cite{sutherland2015error,sriperumbudur2015optimal}, kernel-ridge guarantees \cite{avron2017rffkrr}, generalization bounds \cite{rudi2017generalization}, quadrature connections \cite{bach2017equivalence}, unified risk analyses \cite{li2019unified}, and random-matrix characterizations of high-dimensional random-feature models \cite{liao2020rmt}. A TPAMI survey reviews this broader area \cite{liu2022survey}. Recent data-aware random-feature attention further adapts the projection distribution to anisotropic pretrained queries and keys to reduce Monte Carlo variance \cite{farzam2026dataaware}. Our finite-feature term is related to this RFF approximation-noise problem. The structural term is distinct: it is determined by off-target kernel overlap among stored keys and remains even when the kernel itself is evaluated exactly.

\subsection{Representation geometry, whitening, and similarity structure}
Learned representations can differ substantially in anisotropy, spectral concentration, and pairwise similarity structure \cite{bengio2013representation}. Decorrelation has been used as an explicit regularizer in supervised networks \cite{cogswell2016decov} and as a central ingredient of self-supervised methods, including whitening-based learning \cite{ermolov2021whitening}, Barlow Twins \cite{zbontar2021barlow}, VICReg \cite{bardes2022vicreg}, and feature-decorrelation analyses \cite{hua2021decorrelation}. Contrastive and non-contrastive objectives such as SimCLR and BYOL provide distinct representation-learning regimes \cite{chen2020simclr,grill2020byol}; subsequent work has explicitly analyzed alignment, uniformity, whitening, and dimensional collapse \cite{wang2020alignment,ermolov2021whitening,jing2022collapse}. DINOv2 \cite{oquab2024dinov2} and CLIP \cite{radford2021clip} provide additional self-supervised and vision-language representation families used in our evaluation. Neural and dimensional collapse give further examples of geometry changing independently of nominal dimension \cite{papyan2020collapse,jing2022collapse}. Effective rank formalizes continuous dimensionality \cite{roy2007effectiverank} and has been used to assess self-supervised representations \cite{garrido2023rankme}. Representation-comparison methods such as SVCCA \cite{raghu2017svcca} and centered kernel alignment (CKA) \cite{kornblith2019similarity} use subspace or Gram/kernel structure rather than ambient dimension alone; spectral analyses of self-supervised learning likewise emphasize pairwise relations and task alignment \cite{balestriero2022spectral}. In our setting, these geometric descriptors motivate the analysis but do not substitute for the memory-specific quantity $\sum_{j\neq i}K_{ij}^2$ or, with correlated values, the joint key--value geometry entering the retrieval equation.

\section{Theory}

\subsection{Finite-feature Hebbian memory}

Given stored key--value pairs $\{(k_i,v_i)\}_{i=1}^{N}$ with $v_i\in\mathbb{R}^{d_v}$, let $\phi_R(k)\in\mathbb{R}^{R}$ be a unit-normalized random feature map and define
\[
\hat K_{ij}=\phi_R(k_i)^\top\phi_R(k_j).
\]
For the idealized unbiased construction used in the derivation,
\[
\mathbb{E}[\hat K_{ij}]=K(k_i,k_j),
\]
for a chosen normalized positive-definite kernel $K$ with $K(k,k)=1$; we write $K_{ij}=K(k_i,k_j)$. We additionally assume fixed-pair consistency, $\hat K_{ij}\to K_{ij}$ in probability as $R\to\infty$; for the i.i.d. RFF constructions considered here, the convergence is almost sure. Section~\ref{sec:kernel-rff} states how the separately row-normalized random-phase implementation differs from exact unbiasedness at finite $R$ while retaining the same large-$R$ limit.

Treating $\phi_R(k_j)$ and $v_j$ as column vectors, the Hebbian memory is
\[
M=\sum_{j=1}^{N}\phi_R(k_j)v_j^\top\in\mathbb{R}^{R\times d_v}.
\]
For the exact-key query $q=k_i$, the retrieved value is therefore
\[
\hat v_i=M^\top\phi_R(k_i)
=\sum_{j=1}^{N}\hat K_{ij}v_j
\in\mathbb{R}^{d_v}.
\]
We evaluate the normalized retrieval cosine
\[
C_i(R)=\frac{\langle \hat v_i,v_i\rangle}{\|\hat v_i\|\,\|v_i\|}.
\]

\subsection{Random-value capacity law}

A convenient random-value reference model is $v_j\overset{\mathrm{i.i.d.}}{\sim}\mathrm{Unif}(\mathbb{S}^{d_v-1})$, for which $\mathbb{E}[v_j^\top v_\ell]=0$ for $j\neq\ell$ and off-diagonal inner products concentrate around zero as $d_v$ grows. The mutually orthogonal unit-value model gives the corresponding exact algebraic idealization. In either regime, the deterministic key-overlap term is
\[
I_{\mathrm{struct},i}=\sum_{j\neq i}K_{ij}^{2}.
\]
Intuitively, this term measures how strongly the other stored items resemble item $i$ under the chosen kernel.
For an unbiased finite-feature estimator,
\[
\mathbb{E}[\hat K_{ij}^{2}]=K_{ij}^{2}+\operatorname{Var}(\hat K_{ij}).
\]
Before the weak-overlap simplification, the concentrated cross-talk approximation therefore has the more general form
\begin{equation}
C_i(R)\approx\left[1+I_{\mathrm{struct},i}+\sum_{j\neq i}\operatorname{Var}(\hat K_{ij})\right]^{-1/2}.
\label{eq:general-keylaw}
\end{equation}
For paired sine/cosine RFFs with total feature dimension $R$ and a normalized shift-invariant kernel $K(x,z)=\kappa(x-z)$, the exact pairwise variance is
\[
\operatorname{Var}(\hat K_{ij})=\frac{1+\kappa(2\Delta_{ij})-2K_{ij}^{2}}{R},\qquad \Delta_{ij}=k_i-k_j,
\]
and for the RBF kernel this becomes $(1-K_{ij}^{2})^{2}/R$ \cite{sutherland2015error}. In the weak-overlap regime used to obtain the simple closed form, most off-target $K_{ij}^{2}$ are small, so the summed variance is approximately $(N-1)/R$. This yields
\begin{equation}
\boxed{C_i(R)\approx\left[1+\frac{N-1}{R}+\sum_{j\neq i}K_{ij}^{2}\right]^{-1/2}.}
\label{eq:keylaw}
\end{equation}
Thus the simple law separates the leading finite-feature and structural terms,
\[
I_{\mathrm{total},i}\approx\underbrace{\frac{N-1}{R}}_{\text{finite-feature}}+\underbrace{\sum_{j\neq i}K_{ij}^2}_{\text{structural}}.
\]
As $R\to\infty$, consistency removes the finite-feature term and gives the key-only mean-field ceiling
\begin{equation}
\boxed{C_{i,\infty}\approx\left[1+\sum_{j\neq i}K_{ij}^{2}\right]^{-1/2}.}
\label{eq:keyceiling}
\end{equation}
This expression is exact for mutually orthogonal unit values (Appendix~\ref{app:keyproof}) and is the corresponding concentration approximation for finite-dimensional isotropic random values. Increasing $R$ reduces finite-feature interference but cannot remove the structural ceiling set by key geometry.

\subsection{Target-cosine capacity-boundary prediction}

For a target cosine $C^\star$, a memory-design boundary is obtained by solving
\[
\overline{C(R/N)}=C^\star
\]
for $R/N$, where the overline denotes the average over stored items. When structural interference is negligible, Eq.~\eqref{eq:keylaw} gives the finite-$N$ relation
\[
\frac{R}{N}\approx\frac{1-1/N}{1/(C^\star)^2-1},
\]
which approaches $[1/(C^\star)^2-1]^{-1}$ as $N\to\infty$. For $C^\star=0.95$, the large-$N$ prediction is $R/N\approx9.26$. With heterogeneous nonzero structural interference, we solve the mean item-wise theoretical cosine numerically (Appendix~\ref{app:keyproof}). In design terms, this boundary converts a measured representation geometry into an explicit memory-sizing estimate for a requested retrieval fidelity, while the infinite-$R$ ceiling identifies targets that cannot be reached by increasing $R$ alone.

\subsection{Joint key--value geometry}

For correlated values, the random-value approximation no longer captures the deterministic retrieval geometry. Define
\[
y_{\infty,i}=\sum_{j=1}^{N}K_{ij}v_j,
\qquad
a_i=\langle y_{\infty,i},v_i\rangle,
\qquad
b_i=\|y_{\infty,i}\|^2.
\]
Let $G_{j\ell}=\langle v_j,v_\ell\rangle$ denote the value Gram matrix. Then
\[
a_i=\sum_j K_{ij}G_{ji},\qquad b_i=\sum_{j,\ell}K_{ij}K_{i\ell}G_{j\ell}.
\]
The infinite-feature retrieval cosine is exactly
\begin{equation}
C_{i,\infty}=\frac{\langle y_{\infty,i},v_i\rangle}{\|y_{\infty,i}\|\,\|v_i\|}=\frac{a_i}{\sqrt{b_i}}\quad\text{for unit values and }b_i>0.
\label{eq:jointinf}
\end{equation}
Writing $\hat v_i=y_{\infty,i}+\epsilon_i$, the finite-$R$ cosine contains numerator noise and numerator--denominator coupling. Equation~\eqref{eq:valuelaw} therefore uses a ratio-of-leading-moments/concentration approximation; it is not an identity for $\mathbb{E}[C_i(R)]$. Under the simple finite-feature reference,
\begin{equation}
\boxed{C_i(R)\approx\frac{a_i}{\sqrt{b_i+(N-1)/R}}.}
\label{eq:valuelaw}
\end{equation}
Equation~\eqref{eq:valuelaw} reduces to the key-only law when values are effectively isotropic and the value Gram cross terms vanish.

\subsection{Covariance-aware finite-$R$ correction}\label{sec:cov-correction}

For paired sine/cosine RFFs with $R=2m$, define $D_i$ through the exact second-moment identity
\[
\mathbb{E}\|\epsilon_i\|^2=\frac{D_i}{m}=\frac{2D_i}{R}.
\]
Using this exact noise energy inside the same mean-cosine approximation gives
\begin{equation}
\boxed{C_i^{\mathrm{cov}}(R)\approx\frac{a_i}{\sqrt{b_i+2D_i/R}}.}
\label{eq:covlaw}
\end{equation}
For approximately orthogonal random values under the RBF kernel,
\[
D_i\approx\frac{1}{2}\sum_{j\neq i}(1-K_{ij}^{2})^{2}\approx\frac{N-1}{2},
\]
only when off-target overlaps are predominantly small, recovering Eq.~\eqref{eq:valuelaw}. With strongly correlated aligned real values, off-diagonal covariance terms can make $D_i$ substantially larger.

Equation~\eqref{eq:covlaw} is not a universal replacement for Eq.~\eqref{eq:valuelaw}. In the controls, it improves aligned real-value predictions but can overcorrect permuted values. We therefore use it only for close design comparisons where candidates have similar absolute performance but different finite-$R$ noise.

\section{Experimental Setup}

\subsection{Synthetic key geometries}

We evaluate i.i.d./isotropic keys and controlled departures from isotropy:
\begin{itemize}
    \item clustered keys, with within-cluster correlation controlled by $\rho$;
    \item exact low-rank keys;
    \item anisotropic keys with decaying spectrum.
\end{itemize}
These experiments isolate structural interference while varying $R/N$.

\subsection{Real visual representations}\label{sec:real-reps}

We use the test splits of CIFAR-100 and CIFAR-10 \cite{krizhevsky2009cifar} and the STL-10 test split \cite{coates2011stl}. The original validation uses ImageNet-pretrained ResNet50 \cite{he2016resnet} and ViT-B/16 \cite{dosovitskiy2021vit}. Images are resized to 256 pixels, center-cropped to $224\times224$, and normalized with ImageNet statistics before feature extraction. The CIFAR-100 geometry intervention uses $N=4096$ images. Capacity-boundary experiments use at most $N=2048$ images per CIFAR-10/STL-10 dataset/backbone condition, selected using subset seed 5207. Throughout the paper, ``unfitted-condition'' prediction means that no empirical coefficient or retrieval curve from that condition is fitted; the condition's key geometry (and, for the joint theory, its value geometry) is measured and supplied to the analytic prediction.

To evaluate the same analytic law across additional pretraining paradigms, we extract DINOv2 ViT-B/14 features \cite{oquab2024dinov2} and CLIP ViT-B/16 features with OpenAI-pretrained weights \cite{radford2021clip} on CIFAR-10 and STL-10. DINOv2 uses resize-256/center-crop-224 with ImageNet normalization; CLIP uses its standard 224-pixel preprocessing and CLIP normalization. The resulting DINOv2 and CLIP embeddings have dimensions 768 and 512, respectively, and are $\ell_2$-normalized before memory analysis. The key-only DINOv2/CLIP extension uses $N=2048$ images per dataset/representation condition; the joint key--value controls use $N=1024$ aligned image pairs in each direction.

For each embedding matrix we consider:
\begin{enumerate}
    \item \textbf{RAW}: row-normalized original embeddings;
    \item \textbf{CENTERED}: dataset-mean subtraction followed by row normalization;
    \item \textbf{WHITENED}: PCA whitening followed by row normalization, using 256 retained components where specified.
\end{enumerate}
These transformations act on the representation while leaving the underlying image set unchanged. These geometry interventions can change how distinctly cases are represented in memory, with potential consequences for cross-modal retrieval and case-based reasoning. Centering and whitening statistics are estimated separately for each sampled dataset/backbone representation. Whitening uses explicit mean subtraction followed by \texttt{torch.pca\_lowrank} with $q=\min(256,N-1,d)$, four power iterations, and no internal recentering. PCA scores are scaled by $\sqrt{N-1}/\max(S,10^{-8})$ and then row-normalized; the same sampled observations are used to estimate the transform and to evaluate their memory geometry because the experiment is an intervention on the representation itself rather than a learned downstream predictor.

\subsection{Public and independent paired medical stress tests}\label{sec:clinical}

Our primary medical stress test uses the public SynthRAD2023 Task-1 MRI-to-CT training data \cite{thummerer2023synthrad}. The released training split contains 180 brain and 180 pelvis MRI--CT pairs. We use the released paired data directly, without any additional learned or external registration step. The dataset provides paired planning CT and registered MRI volumes; masks supplied with the challenge are not used here. We load each pair in canonical orientation and, when the voxel grid or affine differs, resample MRI onto the CT grid for slice-wise paired analysis. Foreground-valid CT slices contain at least 3\% finite pixels above $-300$ HU; the middle valid axial index defines the center, and the nearest valid slices at offsets $\{-4,-2,0,+2,+4\}$ are used. The same indices are used for MRI. CT slices are represented with the three fixed Hounsfield-unit windows $(40,80)$, $(100,400)$, and $(600,2800)$ (center, width). MRI slices are independently normalized between their 1st and 99th intensity percentiles and repeated as three grayscale channels. DINOv2 and CLIP embeddings are then extracted with the encoder-specific preprocessing described above.

The primary SynthRAD2023 geometry test is matched-$N$ within each anatomical cohort. For each of 100 random constructions, the \emph{dispersed} condition stores one center slice from 120 distinct patients, whereas the \emph{clustered} condition stores five adjacent slices from 24 patients selected from that same 120-patient pool; both memories therefore contain exactly $N=120$ items. The RBF scale is estimated only from the dispersed reference set and then held fixed for its paired clustered condition. Across the 100 random memory constructions, we summarize paired changes in mean structural interference and mean item-wise $C_\infty$ using their mean, median, empirical construction quantiles, and fraction in the expected direction. We additionally report 95\% bootstrap intervals for the mean construction-level contrast using 100,000 bootstrap resamples of these 100 construction effects. These intervals quantify construction-level uncertainty conditional on the observed cohort and are not interpreted as patient-population confidence intervals. A separate finite-$R$ validation uses five matched-$N$ constructions per cohort under RAW, CENTERED, and WHITENED preprocessing, $R/N\in\{2,4,8,16\}$, and RFF seeds 0 and 1. The public cross-modal correlated-value test uses five patient-level subsets of $N=120$ in both MRI$\rightarrow$CT and CT$\rightarrow$MRI directions with aligned, permuted, and random values.

As an independent NCCT--CTA cohort evaluation, we evaluate 445 paired brain NCCT and CTA examinations obtained from the AcT (Alteplase compared to Tenecteplase) randomized controlled trial. The study was approved by the Conjoint Health Research Ethics Board of the University of Calgary (REB18-1101) and by the local ethics boards of participating sites, with written informed consent provided by patients or authorized surrogates as applicable. All analyses were performed on de-identified data in accordance with institutional data-governance requirements. Because of participant privacy and data-use agreements, the clinical imaging data cannot be publicly released. The paired NCCT and CTA volumes share the same image grid and affine. Five axial slices are selected with the same foreground-valid, center, and two-slice-spacing rule used for SynthRAD2023 and encoded with the same three CT windows. The final matched-$N$ construction experiment uses 100 random constructions with $N=400$: 400 examination-level center slices versus five adjacent slices from 80 examinations selected from the same pool. The finite-$R$ confirmation uses five constructions spanning RAW/CENTERED/WHITENED geometry and $R/N\in\{2,4,8,16\}$, while the correlated-value confirmation uses five $N=256$ subsets in both NCCT$\rightarrow$CTA and CTA$\rightarrow$NCCT directions. Patient-level clinical images and identifiers are not redistributed.

\subsection{Kernel and random features}\label{sec:kernel-rff}
All key analyses use the RBF kernel
\begin{equation}
K(x,z)=\exp\!\left[-\frac{\gamma^2}{2}\|x-z\|_2^2\right]
\label{eq:rbfkernel}
\end{equation}
on row-normalized representations. The automatic scale uses the same median-distance rule throughout. For the CIFAR-100 geometry intervention we use 200,000 sampled off-diagonal pairs; the broad ResNet/ViT and DINOv2/CLIP validations use 150,000 sampled pairs; and the subset crossover calculation uses 100,000. Sampling uses seed 5207. In the 100-construction medical matched-$N$ analyses, all off-diagonal pairs of the dispersed reference subset are used ($N=120$ for each SynthRAD2023 cohort and $N=400$ for the independent NCCT--CTA cohort). In each case,
\begin{equation}
\gamma_0=\sqrt{\frac{2\log 100}{\operatorname{median}\|x_i-x_j\|_2^2}},
\end{equation}
so the median sampled kernel similarity is approximately 0.01. In the RAW/CENTERED/WHITENED intervention of Sec.~\ref{sec:real-reps}, $\gamma_0$ is estimated from RAW features and held fixed across the three transformations. In the broader representation/kernel sweeps and capacity-boundary experiments, $\gamma_0$ is estimated separately for each representation mode and evaluated at multipliers 0.7, 1.0, and 1.4.

The broad capacity validations use random-phase cosine RFFs,
\begin{equation}
z(x)=\sqrt{2/R}\cos(x^\top W+b),
\end{equation}
with columns of $W$ sampled from $\mathcal N(0,\gamma^2 I)$ and $b_r\sim\mathrm{Unif}[0,2\pi)$, followed by row normalization of $z(x)$. Row normalization enforces unit self-inner-products but makes $\mathbb E[z(x)^\top z(z)] = K(x,z)$ only approximate at finite $R$. Before row normalization, the random-phase Monte Carlo inner product and both feature norms satisfy a strong-law limit; hence the row-normalized inner product remains consistent for every fixed pair as $R\to\infty$. Accordingly, the exact paired-RFF variance formula in Sec.~3.2 is not asserted for this separately normalized feature map: Eq.~\eqref{eq:keylaw} is used as a leading-order prediction and is evaluated empirically. The paired construction used for the covariance analysis has the explicit second moments derived in the Appendix. Unless otherwise stated, $R/N\in\{2,4,8,16\}$ and empirical results are averaged over RFF seeds 0 and 1. The headline boundary evaluation is an explicit exception: it uses discovery seeds 100 and 101 to locate fixed-grid brackets and confirmation seeds 0--19 for the final estimates. Random/isotropic values have dimension 128 and are independently sampled and row-normalized for each seed. The DINOv2/CLIP extension uses the default ratios, seeds, preprocessing modes, and kernel-scale multipliers.

The covariance-aware crossover experiment uses a different but explicitly controlled feature map: paired sine/cosine RFFs with $R=2m$. This paired construction is used because its finite-$R$ second moments yield the coefficient $D_i$ in Sec.~\ref{sec:cov-correction}. The paper therefore distinguishes the random-phase normalized RFF used for the broad capacity validation from the paired RFF used for the covariance-sensitive crossover analysis; the latter is not treated as an interchangeable baseline.
\subsection{Real correlated values and alignment controls}\label{sec:value-controls}

For paired visual memories, one backbone provides keys and the other provides values for the same image. We compare:
\begin{itemize}
    \item \textbf{aligned real values}: the original image-wise pairing;
    \item \textbf{permuted real values}: value marginals and correlations preserved, image-wise key--value alignment disrupted;
    \item \textbf{random values}: isotropic random values.
\end{itemize}
The permutation control preserves the multiset of values and its marginal Gram geometry while disrupting image-wise alignment with key-kernel neighborhoods. The finite-$R$ theory-agreement experiments use one fixed permutation per key--value pair, generated from seed $5207+31091$ and shared across preprocessing, kernel scales, and memory budgets; random-value controls use seed $5207+72113$. To assess sensitivity to the particular permutation, we additionally repeat the infinite-$R$ DINOv2--CLIP alignment control with 200 independent permutations for each of the 24 dataset$\times$direction$\times$preprocessing$\times$kernel-scale conditions. For pair index $p$, permutations are generated with NumPy \texttt{default\_rng} seed $832019+100003p$; the same permutation bank is shared across preprocessing and $\gamma$ within a pair. We report the mean and central 95\% empirical permutation interval of $\Delta C_\infty=C_\infty^{\rm aligned}-C_\infty^{\rm permuted}$, the fraction of permutations with $\Delta C_\infty>0$, and the one-sided Monte Carlo permutation $p$-value $(1+n_{\rm perm\ge aligned})/(B+1)$ with $B=200$.

\subsection{Evaluation}\label{sec:evaluation}

We report exact-key retrieval cosine, theory--empirical MAE/RMSE/correlation, target-cosine residual error, interpolated boundary-location error in $R/N$, empirical-oracle regret for design selection, RAW-to-WHITENED crossover error, repeated-permutation uncertainty for the alignment control, and paired matched-$N$ changes in $\Istruct$ and mean item-wise $C_\infty$ for the public and independent medical clustering stress tests. Top-1 recall, where shown in auxiliary outputs, is the fraction of queries for which the retrieved normalized vector has maximum cosine similarity with its own stored value among all $N$ values. For the headline capacity-boundary evaluation, empirical retrieval is evaluated on a fixed $R/N$ grid shared across conditions rather than at theory-centered probes. Two discovery RFF seeds are used only to identify empirical bracketing intervals on this fixed grid, and 20 independent confirmation seeds are then used to estimate the final boundary by linear interpolation within the discovered brackets. The grid is extended below $R/N=2$ for the low-budget region, while retaining the same grid for all conditions in that region. The final headline summary contains 38 CIFAR-10/STL-10 conditions whose target crossings are bracketed within the evaluated budget; one additional condition lies beyond that budget and is excluded. For the complementary target-cosine residual, empirical retrieval is evaluated at the theory-predicted boundary using the same 20 confirmation seeds. For the crossover analysis, each dataset$\times$direction group contains five independently sampled subsets of size 1024. Subset $s$ uses sampling seed $5207+1009s$, yielding 20 subset-level crossovers in total. Uncertainty in the absolute-error reduction is estimated by a block bootstrap that resamples subset IDs within each dataset while preserving the paired forward/reverse directions in each dataset--subset block.

\section{Results}

\subsection{Controlled key geometries}

Across clustered, low-rank, and anisotropic keys, Eq.~\eqref{eq:keylaw} gives small prediction errors relative to the observed retrieval cosine. Clustered-key experiments achieved MAE $0.000626$, RMSE $0.000885$, and $r=0.999992$. Exact low-rank experiments achieved MAE $0.000683$ and $r=0.999777$, while anisotropic experiments achieved MAE $0.000445$ and $r=0.999930$.

The controlled geometries separate finite-feature and structural interference. For example, sufficiently correlated clustered keys retain a low infinite-$R$ ceiling even as $R$ grows, whereas geometries with small kernel overlap approach the finite-feature prediction $[1+(N-1)/R]^{-1/2}$. Likewise, rank-8 keys exhibit a lower structural ceiling, while rank-32 and higher approach the isotropic reference.

\begin{table*}[!t]
\centering
\footnotesize
\caption{Summary of theory--empirical agreement across the main validation stages. The boundary-location row is measured in units of $R/N$; the other prediction-error rows are cosine errors.}\label{tab:master}
\begin{tabular}{lrrr}
\toprule
Experiment & MAE & RMSE & $r$\\
\midrule
Clustered synthetic keys & 0.000626 & 0.000885 & 0.999992\\
Low-rank synthetic keys & 0.000683 & 0.001126 & 0.999777\\
Anisotropic synthetic keys & 0.000445 & 0.000612 & 0.999930\\
ResNet/ViT multi-condition validation & 0.003879 & 0.004838 & 0.999969\\
DINOv2/CLIP extension & 0.000878 & 0.001308 & 0.999992\\
Public SynthRAD2023 finite-$R$ validation & 0.002132 & 0.002868 & 0.999274\\
Independent NCCT/CTA finite-$R$ validation & 0.002031 & 0.002659 & 0.999923\\
CIFAR-100 target-cosine residual & 0.001183 & 0.002095 & 0.999577\\
CIFAR-10/STL-10 target-cosine residual (20 seeds) & 0.000495 & 0.000531 & 0.999997\\
CIFAR-10/STL-10 fixed-grid boundary ($R/N$, 20 seeds) & 0.099807 & 0.199290 & 0.999894\\
Real correlated values, key-only theory & 0.113460 & 0.152437 & 0.865286\\
Real correlated values, joint theory & 0.004209 & 0.007842 & 0.999500\\
Joint controls, joint key--value theory & 0.001828 & -- & 0.999863\\
DINOv2/CLIP joint controls & 0.002926 & 0.006346 & 0.999712\\
\bottomrule
\end{tabular}
\end{table*}

\begin{figure}[t]
\centering
\includegraphics[width=\columnwidth]{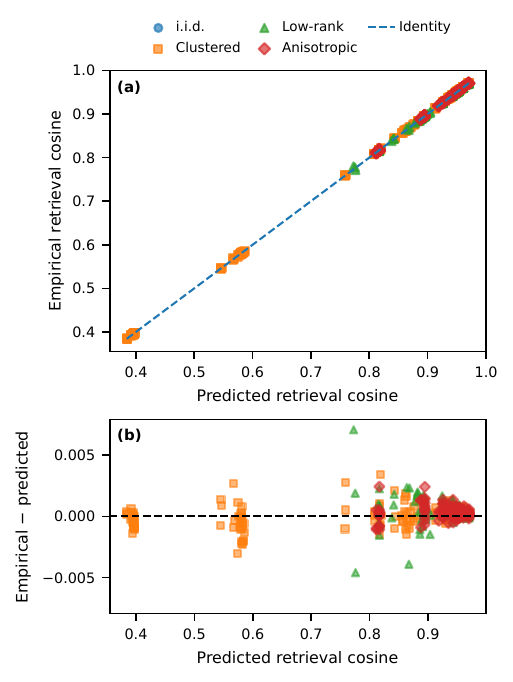}
\caption{Synthetic theory validation over 448 i.i.d., clustered, low-rank, and anisotropic conditions. (a) Predicted versus empirical retrieval cosine; the dashed line is the identity line. (b) Residual $C_{\rm empirical}-C_{\rm predicted}$ on an expanded vertical scale. The overall MAE is $5.79\times10^{-4}$; the residual range is $[-0.00460,\,0.00708]$.}
\label{fig:synthetic}
\end{figure}

\subsection{Geometry interventions on real embeddings}

Raw ResNet50 CIFAR-100 embeddings have nominal dimension 2048 but effective rank about 6.5, with RMS cosine 0.392. The mean item-wise structural interference is 1.658, while the separately averaged item-wise infinite-$R$ ceiling is 0.652. Centering increases effective rank to about 66.5, reduces mean structural interference to 0.0498, and increases the mean item-wise ceiling to 0.979. PCA whitening to 256 dimensions increases effective rank to about 252.7, reduces mean structural interference to 0.00835, and increases the mean item-wise ceiling to 0.996. Because $C_{i,\infty}=(1+I_{\mathrm{struct},i})^{-1/2}$ is nonlinear, the reported mean ceiling is not obtained by substituting the mean structural interference into Eq.~\eqref{eq:keyceiling}.

In this intervention, reducing the nominal dimension from 2048 to 256 is accompanied by higher capacity because the kernelized geometry is improved; ambient dimension alone does not account for the change. Practically, this example shows that increasing representation size is not necessarily the effective response to memory interference: a lower-dimensional but better-separated representation can require less memory to achieve higher retrieval fidelity.

\begin{table*}[!t]
\centering
\footnotesize
\caption{Representative real-embedding geometry statistics. DINOv2/CLIP rows use CIFAR-10 at the base kernel scale ($\gamma\times1.0$). Structural interference and the infinite-$R$ ceiling are reported as separate item-wise averages; because the ceiling is nonlinear in $I_{\mathrm{struct},i}$, mean $C_{i,\infty}$ is not obtained by substituting mean $I_{\mathrm{struct},i}$ into Eq.~\eqref{eq:keyceiling}.}
\label{tab:geometry}
\begin{tabular}{lrrrrr}
\toprule
Backbone / mode & Dim. & Eff.\ rank & RMS cos. & mean $I_{\rm struct,i}$ & mean $C_{i,\infty}$\\
\midrule
ResNet50 RAW & 2048 & 6.52 & 0.392 & 1.658 & 0.652\\
ResNet50 CENTERED & 2048 & 66.50 & 0.122 & 0.0498 & 0.979\\
ResNet50 WHITENED & 256 & 252.68 & 0.061 & 0.00835 & 0.996\\
ViT-B/16 RAW & 768 & 32.98 & 0.173 & 1.354 & 0.680\\
ViT-B/16 CENTERED & 768 & 98.74 & 0.099 & 0.436 & 0.850\\
ViT-B/16 WHITENED & 256 & 253.70 & 0.061 & 0.187 & 0.920\\
DINOv2 RAW & 768 & 61.42 & 0.126 & 0.732 & 0.779\\
DINOv2 WHITENED & 256 & 252.43 & 0.059 & 0.268 & 0.889\\
CLIP RAW & 512 & 1.44 & 0.834 & 3.858 & 0.567\\
CLIP WHITENED & 256 & 246.77 & 0.060 & 0.252 & 0.894\\
\bottomrule
\end{tabular}
\end{table*}

Effective rank is also insufficient to identify the capacity ceiling. Both whitened representations have effective rank near 253, yet their structural interference and capacity ceilings differ.

\begin{figure}[t]
\centering
\includegraphics[width=\columnwidth]{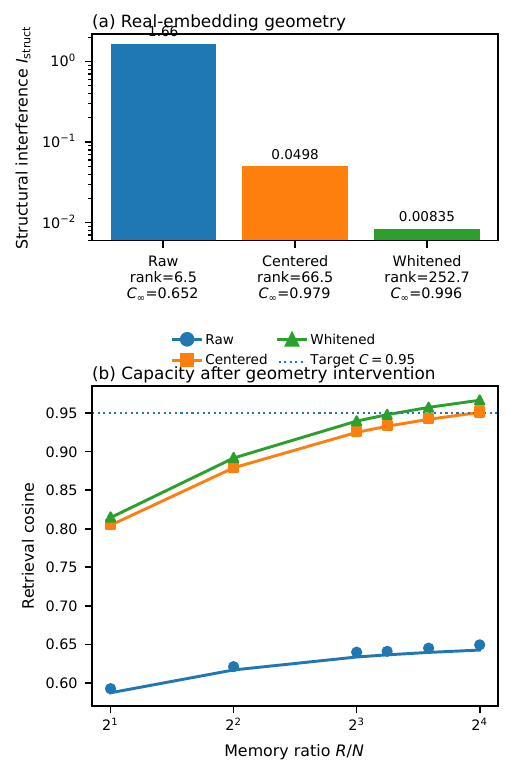}
\caption{Geometry intervention on real CIFAR-100 ResNet50 embeddings. (a) Mean structural interference decreases from 1.658 (RAW) to 0.0498 (CENTERED) and 0.00835 (WHITENED), while the separately averaged item-wise infinite-$R$ ceiling increases; whitening retains 256 components. (b) Lines show theoretical predictions, and markers show empirical retrieval for the same images under the three preprocessing choices.}
\label{fig:geometry}
\end{figure}

\subsection{Prediction across real representations and kernel scales}

Across backbone, preprocessing, kernel scale, and $R/N$, the key-only theory achieves MAE $0.003879$, RMSE $0.004838$, and $r=0.999969$ on the seed-averaged conditions. A fitted diagnostic line has slope 1.0065 and intercept $-0.00765$; these coefficients are descriptive only and are not used by the theory.

As a direct control, we remove the structural term and retain only the finite-feature prediction $[1+(N-1)/R]^{-1/2}$. Across the same 72 seed-averaged real-representation conditions, this finite-feature-only baseline has MAE $0.289832$, compared with $0.003879$ for the geometry-aware law, corresponding to a 98.66\% reduction in MAE. Within this model, including the structural term is necessary to account for the variation across these real-representation conditions.

Changing the kernel scale provides a counterexample to a rank-only explanation. ResNet50 RAW at $\gamma\times1.4$ retains effective rank about 6.5 but has structural interference of only 0.00589 and $C_\infty\approx0.997$. Conversely, whitened representations at $\gamma\times0.7$ have effective rank near 253 but structural interference around 23--24 and $C_\infty\approx0.20$. These comparisons show that effective rank alone does not determine retrieval; the kernel-transformed pairwise geometry is required by the capacity law.

The DINOv2/CLIP representation-family extension applies the same law to additional representation geometries without refitting empirical coefficients. Across 144 seed-averaged conditions spanning CIFAR-10/STL-10, DINOv2/CLIP, RAW/CENTERED/WHITENED preprocessing, three kernel scales, and four memory ratios, the geometry-aware law achieves MAE 0.000878, RMSE 0.001308, and $r=0.999992$. The finite-feature-only baseline has MAE 0.319933 on the same conditions. The agreement holds separately for CIFAR-10 CLIP (MAE 0.000825), CIFAR-10 DINOv2 (0.000790), STL-10 CLIP (0.001265), and STL-10 DINOv2 (0.000631). For example, CIFAR-10 CLIP RAW features have effective rank 1.44 at the base kernel scale, whereas whitening raises effective rank to 246.77 and lowers structural interference from 3.858 to 0.252; the same capacity law remains accurate across this change.

\begin{figure}[t]
\centering
\includegraphics[width=\columnwidth]{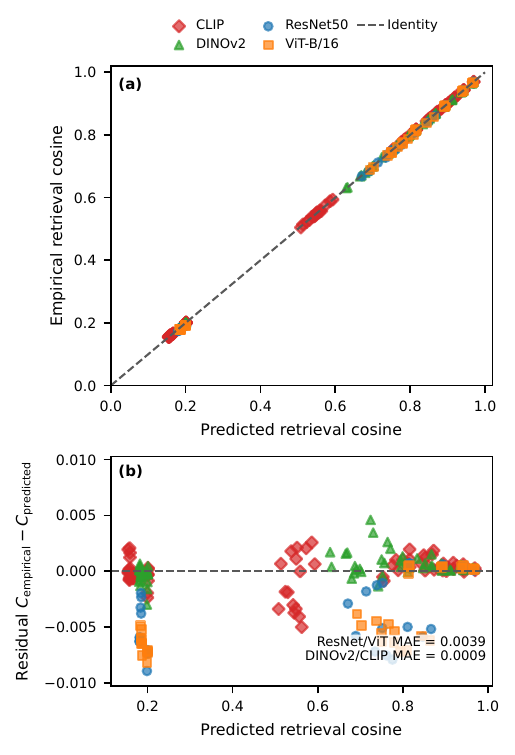}
\caption{Real-representation validation across supervised, self-supervised, and vision-language features. (a) Predicted versus empirical retrieval cosine for ResNet50, ViT-B/16, DINOv2, and CLIP conditions. (b) Signed residuals $C_{\rm empirical}-C_{\rm predicted}$ for the same conditions, showing the error scale directly. The geometry-aware law achieves MAE 0.003879 on the original ResNet/ViT validation and 0.000878 on the DINOv2/CLIP extension; for comparison, the finite-feature-only baseline yields MAE 0.289832 and 0.319933 on the two groups.}
\label{fig:collapse}
\end{figure}

\subsection{Public paired MRI--CT and independent NCCT--CTA cohort evaluation}

The key-only capacity law remains accurate on the public SynthRAD2023 Task-1 training data. Across five matched-$N$ constructions per anatomical cohort, 960 seed-averaged finite-$R$ conditions span brain/pelvis, CT/MRI, DINOv2/CLIP, dispersed/clustered sampling, RAW/CENTERED/WHITENED preprocessing, and four memory ratios. Overall, the agreement has an MAE of 0.002132 and an RMSE of 0.002868, and $r=0.999274$. Thus the same fit-free law remains accurate on a public multi-center paired medical dataset with strong shared anatomy and cross-modal structure.

The 100-construction matched-$N$ experiment compares outcomes after concentrating a fixed-size memory into repeated same-patient anatomical neighborhoods. With $N=120$ in both conditions for each cohort, replacing 120 distinct-patient center slices with 24 patients with five adjacent slices lowers the mean item-wise infinite-feature ceiling in all eight cohort/modality/encoder conditions (Table~\ref{tab:medical}). The conditional bootstrap interval for the mean construction-level contrast $\Delta C_\infty=C_\infty^{\rm clustered}-C_\infty^{\rm dispersed}$ lies strictly below zero in every condition. Seven of eight conditions show the expected $C_\infty$ decrease in all 100 constructions; pelvis CT--CLIP shows the same direction in 96/100. Mean structural interference also increases in all eight conditions, with conditional bootstrap intervals for the mean effect above zero; pelvis CT--CLIP again shows the expected direction in 96/100 constructions. These results mirror the synthetic clustered-key pattern on a fully public anatomical dataset while holding $N$ fixed, although patient diversity and within-patient slice clustering change together in this design.

\begin{table*}[t]
\centering
\footnotesize
\caption{SynthRAD2023 matched-$N$ same-patient clustering over 100 random constructions. Entries are differences between clustered and dispersed means; brackets give 95\% bootstrap intervals for the mean construction-level contrast conditional on the observed cohort. Both conditions contain $N=120$ stored slices.}
\label{tab:medical}
\begin{tabular}{lllrrr}
\toprule
Cohort & Modality & Encoder & $\Delta I_{\rm struct}$ & $\Delta C_\infty$ & $C_\infty\downarrow$\\
\midrule
Brain & CT & CLIP & $+0.273\,[0.255,0.292]$ & $-0.0755\,[-0.0798,-0.0713]$ & 100/100\\
Brain & CT & DINOv2 & $+0.301\,[0.292,0.310]$ & $-0.0997\,[-0.1022,-0.0972]$ & 100/100\\
Brain & MRI & CLIP & $+0.392\,[0.367,0.418]$ & $-0.0865\,[-0.0910,-0.0821]$ & 100/100\\
Brain & MRI & DINOv2 & $+0.305\,[0.292,0.317]$ & $-0.0975\,[-0.1005,-0.0944]$ & 100/100\\
Pelvis & CT & CLIP & $+0.063\,[0.054,0.072]$ & $-0.0225\,[-0.0253,-0.0197]$ & 96/100\\
Pelvis & CT & DINOv2 & $+0.104\,[0.096,0.112]$ & $-0.0382\,[-0.0408,-0.0356]$ & 100/100\\
Pelvis & MRI & CLIP & $+0.200\,[0.193,0.208]$ & $-0.0693\,[-0.0717,-0.0669]$ & 100/100\\
Pelvis & MRI & DINOv2 & $+0.280\,[0.267,0.294]$ & $-0.0866\,[-0.0900,-0.0833]$ & 100/100\\
\bottomrule
\end{tabular}
\end{table*}

The independent 445-pair NCCT--CTA cohort gives a closely matching result. Across 480 seed-averaged finite-$R$ conditions, the law achieves MAE 0.002031, RMSE 0.002659, and $r=0.999923$. In its separate $N=400$, 100-construction matched-$N$ test, mean $\Delta C_\infty$ is negative for all four NCCT/CTA $\times$ DINOv2/CLIP conditions, with conditional bootstrap intervals for the mean construction contrast strictly below zero and the expected direction in 97--100 of 100 constructions. Mean $\Delta I_{\rm struct}$ is positive in all four conditions; construction-level consistency is weaker for CTA--CLIP, where $\Delta I_{\rm struct}>0$ in 60/100 constructions despite a positive conditional interval for the mean. The public and independent cohorts therefore support the same geometry-to-capacity relationship under different paired medical modalities and sampling designs.

\subsection{Capacity-boundary prediction on unfitted conditions}

We solve Eq.~\eqref{eq:keylaw} for the ratio $\lambda_{\rm th}=R/N$ required to reach a requested mean retrieval cosine $C^\star$. This converts the theory from a retrospective description of interference into a prospective memory-sizing calculation. Here ``new'' means that no empirical coefficient or retrieval curve from CIFAR-10/STL-10 is fitted; their measured geometry is supplied to the analytic law. In the earlier CIFAR-100 analysis, 21 reachable targets give a target-cosine residual MAE of 0.001183, RMSE 0.002095, and $r=0.999577$.

For the headline CIFAR-10/STL-10 evaluation with ResNet50 and ViT-B/16, empirical crossings are located on a fixed $R/N$ grid shared across conditions rather than by probes centered on $\lambda_{\rm th}$. Two discovery RFF seeds identify the bracketing interval, after which 20 independent confirmation seeds estimate the empirical boundary by interpolation within that bracket. The final headline summary contains 38 conditions whose crossings are bracketed within the evaluated budget. The resulting boundary-location agreement is
\begin{align*}
\mathrm{MAE}_{R/N}&=0.099807, &
\mathrm{RMSE}_{R/N}&=0.199290,\\
r&=0.999894. &&
\end{align*}
The mean signed error $R/N_{\rm theory}-R/N_{\rm empirical}$ is $+0.014055$, the median absolute error is 0.050977, and the mean relative absolute error is 1.517\%. The maximum absolute error is 0.853894 and occurs for a high-budget condition, where its relative error remains small.

As a complementary diagnostic, empirical retrieval evaluated at $\lambda_{\rm th}$ with the same 20 confirmation seeds gives
\begin{align*}
\mathrm{MAE}_{C}&=0.000495, &
\mathrm{RMSE}_{C}&=0.000531,\\
r&=0.999997,
\end{align*}
with mean signed residual $C_{\rm empirical}-C^\star=+0.000495$, median absolute residual 0.000473, and maximum absolute residual 0.001129. The boundary-location metric tests the predicted memory budget, whereas the target-cosine residual tests how closely empirical retrieval matches the requested target at the theory-predicted budget.

\begin{table}[t]
\centering
\footnotesize
\caption{Representative theory-predicted $R/N$ values for a target cosine $C=0.95$ on CIFAR-10/STL-10 unfitted-condition evaluations.}
\label{tab:holdout}
\begin{tabular}{llr}
\toprule
Dataset & Representation & Predicted $R/N$\\
\midrule
CIFAR-10 & ResNet50 RAW & 10.52\\
CIFAR-10 & ResNet50 CENTERED & 10.54\\
CIFAR-10 & ResNet50 WHITENED & 9.30\\
CIFAR-10 & ViT-B/16 RAW & 13.01\\
CIFAR-10 & ViT-B/16 CENTERED & 12.49\\
CIFAR-10 & ViT-B/16 WHITENED & 9.31\\
STL-10 & ResNet50 RAW & 30.60\\
STL-10 & ResNet50 CENTERED & 28.59\\
STL-10 & ResNet50 WHITENED & 9.67\\
STL-10 & ViT-B/16 WHITENED & 9.59\\
\bottomrule
\end{tabular}
\end{table}

\begin{figure}[!t]
\centering
\includegraphics[width=\columnwidth]{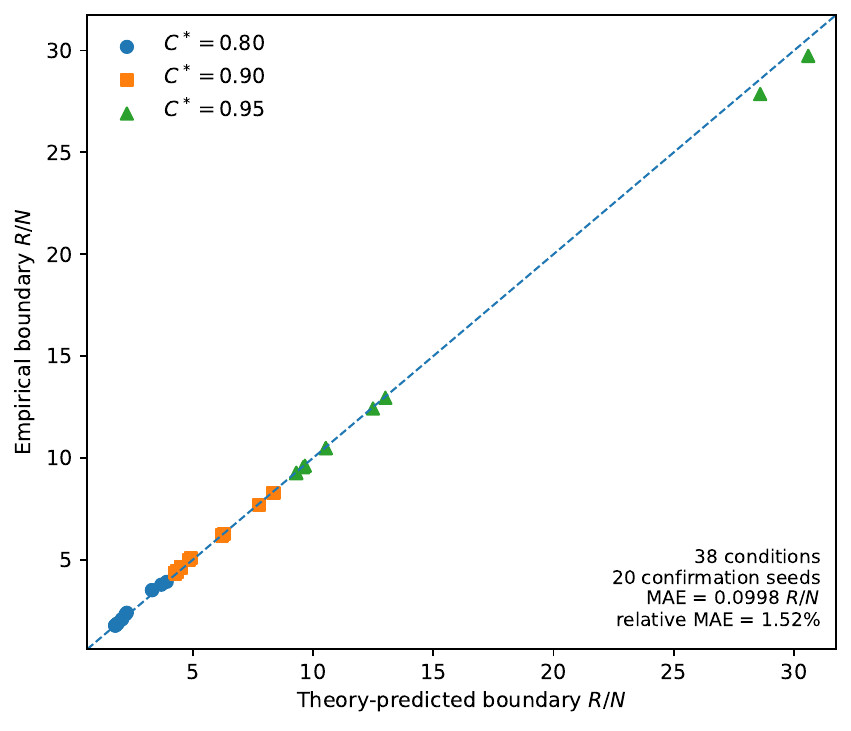}
\caption{Fixed-grid 20-seed target-cosine boundary revalidation on CIFAR-10/STL-10. Theory-predicted versus empirical $R/N$ boundaries for the 38 conditions whose target crossings are bracketed within the evaluated budget. Empirical brackets are located on a fixed grid using discovery seeds and the boundaries are estimated with 20 independent confirmation seeds. The boundary-location MAE is 0.099807 in units of $R/N$, the mean relative absolute error is 1.517\%, and $r=0.999894$.}
\label{fig:boundary}
\end{figure}

\subsection{Joint key--value geometry for correlated values}

When real ViT values are paired with real ResNet keys from the same images, the key-only random-value approximation becomes inadequate. The joint formulation first computes the exact deterministic infinite-$R$ geometry through $y_{\infty,i}$, $a_i$, and $b_i$; only the finite-$R$ perturbation in Eq.~\eqref{eq:valuelaw} is approximate. Across CIFAR-10 and STL-10, the key-only theory has MAE 0.113460, RMSE 0.152437, and $r=0.865286$, whereas the joint formulation has MAE 0.004209, RMSE 0.007842, and $r=0.999500$. Thus the large improvement should be interpreted as identifying the correct deterministic key--value geometry and then showing that a simple finite-$R$ correction remains accurate, rather than as adding a fitted correction to the key-only law.

\begin{table}[t]
\centering
\footnotesize
\caption{Key-only theory versus joint key--value theory for real correlated values.}
\label{tab:valueaware}
\begin{tabular}{lrrr}
\toprule
Setting & MAE & RMSE & $r$\\
\midrule
Key-only theory, all & 0.113460 & 0.152437 & 0.865286\\
Joint key--value, all & 0.004209 & 0.007842 & 0.999500\\
Joint key--value, CIFAR-10 & 0.002878 & -- & 0.999718\\
Joint key--value, STL-10 & 0.005540 & -- & 0.999400\\
\bottomrule
\end{tabular}
\end{table}

The aligned, permuted, and random-value controls separate value correlation from key--value alignment. Across all control conditions, the joint theory has MAE 0.001828 and $r=0.999863$; for random values, the original key-only theory also becomes accurate (MAE 0.000751, $r=0.999992$). Permuting values preserves their marginal Gram geometry but reduces the infinite-$R$ cosine, so the effect depends on how value similarity aligns with neighborhoods induced by the key kernel rather than on marginal value correlation alone.

The DINOv2/CLIP extension reproduces the same pattern in both mapping directions. Across aligned, permuted, and random-value conditions, key-only MAE is 0.115619 ($r=0.693075$), whereas joint-theory MAE is 0.002926 ($r=0.999712$); for aligned real values alone, MAE decreases from 0.170292 to 0.004745. With 200 independent permutations per condition, the mean $\Delta C_\infty=C_\infty^{\rm aligned}-C_\infty^{\rm permuted}$ is positive in all 24 conditions, every central 95\% permutation interval lies above zero, and the aligned ceiling exceeds all 200 permuted realizations in every condition (one-sided Monte Carlo $p=1/201\approx0.00498$). The mean gain across conditions is 0.02991. Directional asymmetry is expected because exchanging keys and values changes both the key kernel and the value Gram geometry.

Public paired MRI--CT data provide a second correlated-value stress test without introducing a fitted term. Across five $N=120$ subsets in both mapping directions, aligned brain CLIP values yield a key-only MAE of 0.1568 versus a joint-theory MAE of 0.01386, and aligned brain DINOv2 values yield 0.09038 versus 0.00888. The same pattern holds in the pelvis: 0.1099 versus 0.01156 for CLIP and 0.06594 versus 0.00729 for DINOv2, corresponding to an MAE reduction of about 89--91\% across the four public aligned-value conditions. For random values, key-only MAE is already 0.0032--0.0057, consistent with the assumptions of Eq.~\eqref{eq:keylaw}. Aligned and permuted SynthRAD2023 conditions have similar joint-theory errors, so this experiment supports the need for joint key--value geometry under correlated cross-modal values rather than a claim of a large subject-wise MRI--CT alignment gain. The independent NCCT--CTA cohort provides the same qualitative pattern: aligned-value MAE decreases from 0.5061 to 0.01447 for CLIP and from 0.2702 to 0.01876 for DINOv2, while random-value key-only errors remain small.

\begin{figure*}[!t]
\centering
\includegraphics[width=0.78\textwidth]{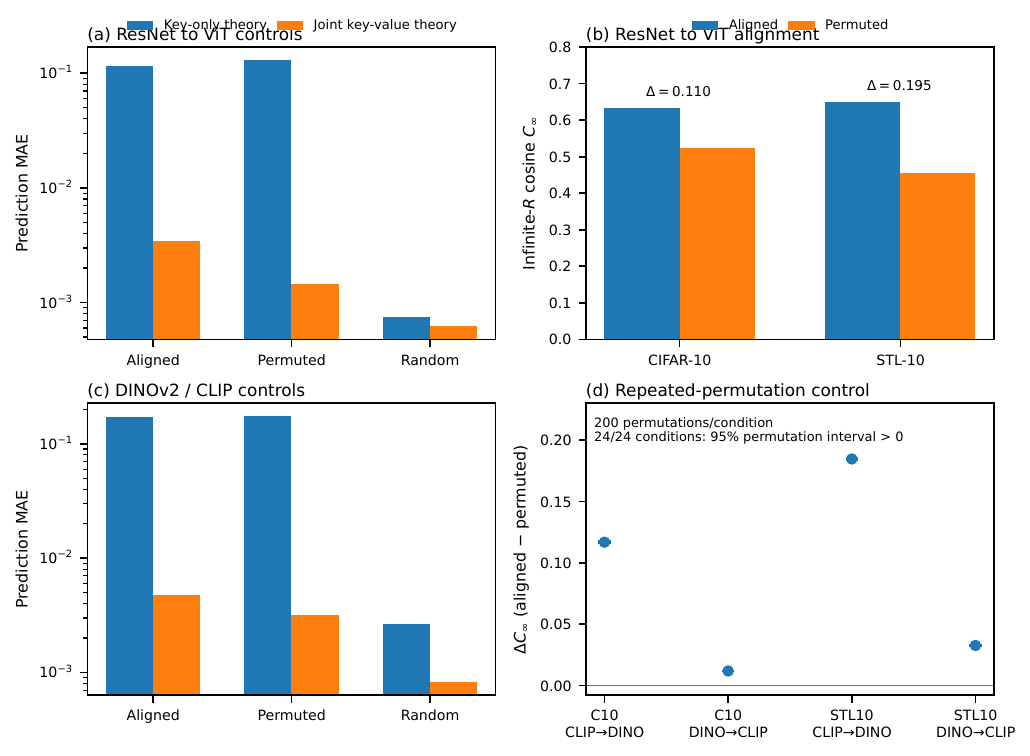}
\caption{Joint key--value controls across representation families. (a, b) Original ResNet$\rightarrow$ViT controls. (c) DINOv2/CLIP prediction error for aligned, permuted, and random values. (d) Repeated-permutation estimates of $\Delta C_\infty=C_\infty^{\rm aligned}-C_\infty^{\rm permuted}$ for representative RAW conditions with $\gamma=1.0$; error bars are central 95\% permutation intervals over 200 permutations. Across all 24 DINOv2/CLIP alignment conditions, the mean $\Delta C_\infty$ and the corresponding permutation interval are strictly positive.}
\label{fig:joint}
\end{figure*}

\subsection{Finite-$R$ covariance is regime dependent}

The simple finite-$R$ correction gives the lowest global error over the tested range of $R/N$. A covariance-denominator correction improves aligned real values but can over-correct permuted values. A second-order delta expansion was also tested and had larger global error, particularly at small $R/N$; it became competitive only at very large $R$. We therefore do not use higher-order delta terms in the main theory.

Accordingly, Eq.~\eqref{eq:covlaw} is used as a targeted design-boundary correction rather than as a replacement for the simpler absolute-capacity approximation.

\subsection{Theory-guided configuration selection}

Using aligned real values, we let theory select among RAW/WHITENED preprocessing and kernel multipliers. This experiment asks whether the same quantities that explain retrieval can also reduce empirical trial-and-error when choosing a representation and kernel for a fixed memory budget. The empirical oracle is the best configuration measured at each memory budget. The key-only theory has mean regret 0.001138; the joint key--value theory has mean regret 0.002568; an arbitrary fixed default (RAW, $\gamma=1.0$) has mean regret 0.271329. Both theory-based selectors have much lower regret than the fixed default, although the joint key--value selector is not uniformly better than the key-only selector for the discrete candidate set.

The empirical optimum changes with memory budget: RAW can be preferable at low budgets, whereas WHITENED becomes preferable at higher budgets. Thus there is no single representation choice that is uniformly preferable across storage constraints; the useful design depends on the available memory budget. This motivates direct prediction of the RAW-to-WHITENED crossover.

\subsection{Covariance-aware crossover prediction}

We summarize subset-level empirical crossovers for two datasets and both backbone directions. Across all 20 subsets, the simple theory overestimates the crossover on average, whereas the covariance-aware theory underestimates it on average. The covariance-aware prediction is closer to the empirical crossover in all 20 subsets and reduces overall MAE by 84.5\%.

\begin{table*}[!t]
\centering
\footnotesize
\caption{Cross-directional RAW-to-WHITENED crossover summary. Lower MAE is better.}
\label{tab:crossover}
\resizebox{0.98\textwidth}{!}{
\begin{tabular}{lrrrrrr}
\toprule
Group & $n$ & Empirical mean & Simple mean & Cov.\ mean & Simple MAE & Cov.\ MAE\\
\midrule
CIFAR-10 R50$\rightarrow$ViT & 5 & 2.7867 & 5.0577 & 2.1620 & 2.2710 & 0.6247\\
STL-10 R50$\rightarrow$ViT & 5 & 2.2956 & 5.1350 & 2.0911 & 2.8394 & 0.2045\\
CIFAR-10 ViT$\rightarrow$R50 & 5 & 3.2725 & 6.7342 & 2.4528 & 3.4617 & 0.8197\\
STL-10 ViT$\rightarrow$R50 & 5 & 2.3525 & 5.9092 & 2.1175 & 3.5567 & 0.2349\\
\midrule
\textbf{ALL} & \textbf{20} & \textbf{2.6768} & \textbf{5.7090} & \textbf{2.2058} & \textbf{3.0322} & \textbf{0.4710}\\
\bottomrule
\end{tabular}}
\end{table*}

The overall absolute-error reduction is 2.561244, corresponding to a relative MAE reduction of 84.468\%. In all groups and overall,
\[
r_{\mathrm{cov\ pred}}
\le
r_{\mathrm{empirical}}
\le
r_{\mathrm{simple\ pred}},
\]
with a bracketing fraction of 1.0. A block bootstrap over the 10 dataset--subset blocks, preserving the paired forward/reverse directions within each block, gives a mean reduction of 2.561244 with a 95\% interval of $[2.415581,\,2.735610]$. The same ordering is observed in both datasets and both mapping directions, although a smaller residual bias remains.

\begin{figure}[!ht]
\centering
\includegraphics[width=\columnwidth]{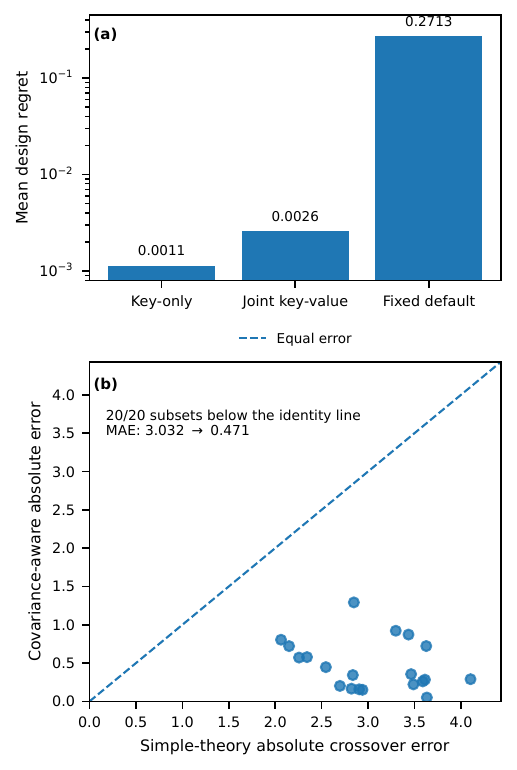}
\caption{Theory-guided design. (a) Mean regret of two theory-based selectors and a fixed RAW default with $\gamma=1.0$. (b) Absolute crossover errors for the simple and covariance-aware predictions over 20 subset-level forward and reverse conditions (10 dataset--subset blocks). Every point lies below the equal-error line; overall MAE decreases from 3.0322 to 0.4710.}
\label{fig:crossover}
\end{figure}

\section{Discussion}

The experiments separate two effects that are often conflated in finite-feature associative memory: approximation noise from the finite random map and deterministic cross-talk induced by stored representation geometry. With correlated values, the relevant deterministic object expands from the key kernel to joint key--value geometry. This separation has a direct design consequence: poor retrieval caused by finite-feature noise calls for more memory, whereas poor retrieval caused by structural overlap calls for a different representation, kernel, or organization of stored items. The main implications are summarized below.

\subsection{Geometry, rather than nominal dimension or rank, controls the structural ceiling}
Our ResNet experiment gives a direct counterexample to dimension-only reasoning: PCA whitening reduces nominal dimension from 2048 to 256 while increasing the mean item-wise infinite-$R$ ceiling from about 0.65 to about 1.0. Effective rank is informative about spectral concentration \cite{garrido2023rankme}, but it is not sufficient either. Representations with similar effective rank can have different structural interference, and changing kernel scale can reverse capacity while leaving rank unchanged. In this compressed random-feature regime, the relevant quantity is therefore the representation geometry as seen through the memory kernel. Consequently, nominal width or embedding dimension alone is not a reliable guide for memory provisioning: representations should be evaluated by the interference they induce under the intended storage rule.

\subsection{The infinite-$R$ ceiling and finite-$R$ noise are distinct}
Structural interference remains even with an exact kernel, whereas finite random-feature noise decreases with increasing $R$. This separation explains why some geometries improve with additional feature budget while others saturate below a target cosine. It also clarifies the operational meaning of the capacity boundary used here: the required $R/N$ depends on both finite-feature noise and the structural ceiling of the measured key set. In practical terms, the theory can indicate whether additional feature dimensions are expected to yield useful improvements in retrieval quality or whether the system has already entered a geometry-limited regime in which extra memory has diminishing benefit.

\subsection{Anatomical clustering recurs across public and independent medical data}
The public SynthRAD2023 experiment extends the geometry test beyond natural-image classification benchmarks with a publicly available paired medical dataset. At matched $N$, concentrating the memory on multiple adjacent slices from fewer subjects lowers mean item-wise $C_\infty$ across brain/pelvis, CT/MRI, and DINOv2/CLIP conditions; an independent NCCT--CTA cohort shows the same pattern. This is consistent with the clustered-key mechanism observed synthetically: same-patient anatomical clustering changes the kernel-overlap distribution without changing nominal embedding dimension or the number of stored items. Because subject diversity and within-patient sampling vary together, these experiments do not isolate the effect of slice adjacency alone. Instead, they test whether stronger similarity among samples from the same patient leads to greater kernel overlap and retrieval interference at fixed \(N\). The resampling intervals describe variability within the observed cohorts rather than uncertainty for new patients. The goal is not to define a modality-specific capacity law, but to test whether the same geometry--interference relationship extends to structured medical-image data. For imaging systems, the practical concern is that large collections can contain many highly similar anatomical observations, so the number of images or feature dimensions alone may overstate how much distinct information a compressed memory can preserve.

\subsection{Correlated values require joint geometry, while design comparisons are more sensitive}
With correlated real values, cross-talk can reinforce or oppose the target. The kernel-weighted mixture $y_{\infty,i}$ makes this dependence explicit, and the aligned/permuted controls show that the relevant object is the alignment between key-kernel neighborhoods and value similarity. The DINOv2--CLIP permutation analysis reduces concern about dependence on a single favorable shuffle. The simple joint finite-$R$ law is accurate over most tested conditions, whereas the covariance-aware correction is useful mainly for closely matched design curves: small systematic errors that barely affect global MAE can still move a RAW-to-WHITENED crossover. This distinction is relevant when theory is used for configuration selection, because the practical question is often not only whether retrieval can be predicted accurately in absolute terms, but whether two competing designs can be ordered correctly at the available memory budget.

\subsection{Practical interpretation for retrieval-oriented AI systems}
In a deployed retrieval system, structural interference means that many stored cases occupy similar neighborhoods under the chosen representation and kernel, so a query can retrieve a mixture of several plausible cases rather than a clean match. This is relevant to tasks such as retrieval of similar prior images, cross-modal image--report or image--image recall, and case-based reasoning, where memory quality depends not only on the number of stored cases but also on how distinctly they are represented. The theory therefore suggests a practical sequence for system design: first measure kernel overlap in the intended representation, then use the predicted capacity curve to determine whether the target retrieval quality is reachable at the available feature budget. If finite-feature noise dominates, increasing $R$ is the natural remedy; if the structural ceiling is already limiting, changing preprocessing, representation, kernel scale, or the composition of the stored set is more appropriate than simply allocating a larger memory.

\subsection{Scope relative to other memory regimes}
Modern Hopfield networks and attention-based memories explicitly retain or parameterize stored patterns and therefore operate in a different storage regime \cite{ramsauer2021hopfield,millidge2022universal}. Direct MLP mappings are also not storage-equivalent comparators for the fixed compressed matrix studied here. Likewise, the negative controls do not support a generic orthogonality advantage: square orthogonal value codebooks are lossless rotations, optimizer-dependent code differences largely collapse under SGD, sparse compositional codes do not outperform Gaussian codes, and learned orthogonal addressing does not consistently beat a noise-trained direct MLP. Our retained claim is narrower: measured kernel geometry predicts interference and target-cosine capacity in this compressed finite-feature regime.

\section{Discussion}
A practical extension is retrieval from noisy or incomplete queries, which are common in clinical settings because repeated or multimodal observations rarely produce identical representations. In such cases, retrieval error would combine query mismatch with the storage interference studied here. This distinction may also matter in applications such as similar-case retrieval or multimodal recall, where failure can arise either from limited memory resources or from overlap among clinically similar representations. Extending the present analysis beyond exact-key retrieval could therefore help separate these sources of error and guide whether to increase memory capacity, change the representation, or improve query matching.

\section{Limitations}

The analysis is restricted to exact-key queries; robustness to corrupted, partial, or shifted cues is not modeled. Extending the framework to noisy or incomplete queries is a natural next step for clinical retrieval settings, where the query representation may differ from any stored case because of acquisition variability, pathology, or missing information. Capacity is defined operationally through target retrieval cosine and feature budget $R/N$, not as an asymptotic pattern-count capacity for other memory mechanisms. The finite-budget derivation is specific to random-feature Hebbian retrieval, and the real-data experiments use pretrained embeddings rather than an end-to-end learned memory. The covariance-aware correction remains approximate and retains residual crossover bias, while the simple finite-$R$ law is less accurate for strongly correlated values at very low $R$, where expectation-of-ratio effects are non-negligible. The medical stress tests use 2-D axial slices and general-purpose encoders rather than 3-D or medically specialized representations, and they evaluate representation geometry and memory interference rather than clinical prediction; therefore, the present results should not be interpreted as evidence that the capacity law directly improves diagnostic or prognostic performance. The public experiment is limited to the SynthRAD2023 Task-1 training split. Additional modalities, 3-D encoders, non-visual data, and end-to-end learned memories would provide more stringent tests of generality.

\section{Conclusion}

For exact-key retrieval in compressed random-feature Hebbian memory, retrieval interference separates into a finite-feature component and a structural component determined by kernel overlap among stored keys. The resulting law predicts target-cosine capacity across synthetic, natural-image, and medical representation geometries, including a public matched-$N$ SynthRAD2023 MRI--CT stress test and an independent NCCT--CTA confirmation. For correlated values, the deterministic retrieval geometry is governed jointly by the key kernel and value Gram matrix, while only the finite-$R$ perturbation is approximated. The same quantities can also estimate the memory budget required for a target retrieval quality, rank candidate representation and kernel configurations, and localize design crossovers. Within this storage regime, the practical value of the theory is therefore not only to describe capacity but also to distinguish when a system is limited by insufficient finite-feature budget from when it is limited by representation geometry itself. Measured kernel geometry is consequently a more informative guide to memory sizing and representation selection than nominal dimension or effective rank alone.

\section*{Data and Code Availability}
The SynthRAD2023 Task-1 training data used for the public medical validation are publicly available from the challenge dataset release \cite{thummerer2023synthrad}. Reproduction code and aggregate source tables for the synthetic, natural-image, SynthRAD2023, and independent cohort analyses will be released publicly upon publication at \url{https://github.com/jianhai-zhang/memory_capacity}. The independent cohort analysis uses paired NCCT--CTA examinations from the AcT (Alteplase compared to Tenecteplase) randomized controlled trial. The study was approved by the Conjoint Health Research Ethics Board of the University of Calgary (REB18-1101) and by local ethics boards at participating sites, with written informed consent provided by patients or authorized surrogates as applicable. Analyses used de-identified data under institutional data-governance requirements. Because of participant privacy and data-use agreements, the clinical imaging data cannot be publicly released.

\appendices

\section{Background and Notation}\label{app:background}

\subsection{Shift-invariant kernels and the two RFF constructions used here}
For a normalized real shift-invariant positive-definite kernel $K(x,z)=\kappa(x-z)$ with $K(x,x)=1$, Bochner's representation implies that $\kappa(\delta)$ can be written as an expectation of Fourier features; RFFs approximate this expectation by Monte Carlo sampling \cite{rahimi2007random,scholkopf2002learning,liu2022survey}. The broad capacity validations use random-phase cosine features followed by row normalization, as specified in Sec.~\ref{sec:kernel-rff}. The covariance analysis uses a paired sine/cosine construction because its second moments admit the clean expression below. For that paired construction, let $R=2m$ and draw independent frequencies $\omega_r$ from the spectral density of the kernel. Define
\begin{equation}
\begin{aligned}
\phi_R(x)=\frac{1}{\sqrt m}\big[&
\cos(\omega_1^\top x),\sin(\omega_1^\top x),\ldots,\\
&\cos(\omega_m^\top x),\sin(\omega_m^\top x)
\big]^\top .
\end{aligned}
\label{eq:app-rff}
\end{equation}
Then
\begin{equation}
\begin{aligned}
\hat K_{ij}
&=\phi_R(k_i)^\top\phi_R(k_j)\\
&=\frac{1}{m}\sum_{r=1}^{m}
\cos\!\left(\omega_r^\top\Delta_{ij}\right),
\qquad \Delta_{ij}=k_i-k_j,
\end{aligned}
\label{eq:app-khat}
\end{equation}
where $\mathbb E[\hat K_{ij}]=K_{ij}$ and $\hat K_{ii}=1$ exactly. By the strong law of large numbers, $\hat K_{ij}\to K_{ij}$ almost surely for every fixed pair. The random-phase cosine estimator used in the broad validation has the same fixed-pair limit before row normalization; its feature norms converge almost surely to one, so the row-normalized inner product is also consistent even though it is not exactly unbiased at finite $R$. Structured and variance-reduced alternatives to this Monte Carlo construction have been studied extensively \cite{le2013fastfood,yang2014qmc,yu2016orf,sutherland2015error,sriperumbudur2015optimal}.

\subsection{Hebbian memory as a kernel-weighted value mixture}
Let $v_j\in\mathbb{R}^{d_v}$ be values represented as column vectors. Stacking a finite-feature address with values gives the Hebbian memory
\begin{equation}
M=\sum_{j=1}^{N}\phi_R(k_j)v_j^\top .
\end{equation}
Querying a stored key $k_i$ yields
\begin{equation}
\hat v_i=M^\top\phi_R(k_i)
=\sum_{j=1}^{N}\hat K_{ij}v_j.
\label{eq:app-retrieval}
\end{equation}
Thus the memory is a randomized kernel smoother in value space. In contrast to explicit key--value memories \cite{weston2015memory,sukhbaatar2015end,ramsauer2021hopfield,millidge2022universal}, the individual pairs are not retained after they are aggregated into $M$.

\subsection{Why effective rank is not enough}
For singular values $\{\sigma_q\}$, one common effective-rank definition is
\begin{equation}
\operatorname{erank}(X)=\exp\!\left(-\sum_q p_q\log p_q\right),
\qquad
p_q=\frac{\sigma_q}{\sum_s\sigma_s},
\label{eq:erank}
\end{equation}
which continuously measures spectral dimensionality \cite{roy2007effectiverank,garrido2023rankme}. Equation~\eqref{eq:erank} depends only on the singular-value spectrum. By contrast, the structural interference $\sum_{j\neq i}K_{ij}^2$ depends on pairwise positions \emph{after} the chosen kernel and kernel scale are applied. Consequently, equal effective rank does not imply equal memory interference. Whitening can alter these pairwise distances because it is generally not an orthogonal transform; this explains why RAW and WHITENED representations can have different memory capacity even when the whitened representation has lower nominal dimension.

\section{Derivation of the Key-Only Capacity Law}\label{app:keyproof}

\subsection{Exact conditional cosine under orthogonal values}
We first expose the interference mechanism in the idealized case of unit, mutually orthogonal values, $\langle v_j,v_\ell\rangle=\delta_{j\ell}$. Because $\hat K_{ii}=1$, Eq.~\eqref{eq:app-retrieval} becomes
\begin{equation}
\hat v_i=v_i+\sum_{j\neq i}\hat K_{ij}v_j.
\end{equation}
The target projection and squared norm are therefore
\begin{equation}
\langle \hat v_i,v_i\rangle=1,
\qquad
\|\hat v_i\|^2=1+\sum_{j\neq i}\hat K_{ij}^2,
\end{equation}
and the conditional retrieval cosine is exactly
\begin{equation}
C_i(R\mid\hat K)
=\left(1+\sum_{j\neq i}\hat K_{ij}^2\right)^{-1/2}.
\label{eq:exact-conditional}
\end{equation}
For high-dimensional random isotropic values, the same expression is a mean-field approximation because off-diagonal value inner products concentrate near zero. More precisely, conditional on coefficients $\{a_j\}_{j\neq i}$ and independent isotropic unit values in dimension $d_v$, let
\[
z_i=\sum_{j\neq i}a_jv_j,\qquad
A_i=\sum_{j\neq i}a_j^2,\qquad
B_i=\sum_{j\neq i}a_j^4.
\]
Then
\begin{equation}
\operatorname{Var}\!\left(v_i^\top z_i\right)=\frac{A_i}{d_v},
\qquad
\operatorname{Var}\!\left(\|z_i\|^2\right)
=\frac{2(A_i^2-B_i)}{d_v}.
\label{eq:finite-value-concentration}
\end{equation}
Thus the random-value deviations are controlled by the total cross-talk energy and $d_v$, rather than by $N/d_v$ alone, and vanish in the concentration limit at fixed $A_i$.

\subsection{Finite-feature and structural terms}
Write $\hat K_{ij}=K_{ij}+\delta_{ij}$ with $\mathbb E[\delta_{ij}]=0$. Then
\begin{equation}
\mathbb E[\hat K_{ij}^2]
=K_{ij}^2+\operatorname{Var}(\hat K_{ij}).
\label{eq:k2split}
\end{equation}
For the paired RFF estimator in Eq.~\eqref{eq:app-khat}, used for the covariance-sensitive analysis, independence across the $m$ frequency draws gives the standard exact second-moment formula \cite{sutherland2015error}
\begin{align}
\operatorname{Var}(\hat K_{ij})
&=\frac{1}{m}\operatorname{Var}_{\omega}
\left[\cos(\omega^\top\Delta_{ij})\right]\\
&=\frac{1}{R}\left[
1+\kappa(2\Delta_{ij})-2K_{ij}^2
\right]
\equiv \frac{\nu_{ij}}{R},
\label{eq:rffvar}
\end{align}
where we used $\cos^2 u=(1+\cos 2u)/2$ and $R=2m$. For the RBF kernel used throughout the experiments,
\[
\kappa(2\Delta_{ij})=K_{ij}^4,
\]
so the paired-RFF variance coefficient simplifies exactly to
\begin{equation}
\boxed{
\nu_{ij}=(1-K_{ij}^2)^2.
}
\label{eq:rbf-nu}
\end{equation}
Combining Eqs.~\eqref{eq:exact-conditional}--\eqref{eq:rffvar} and replacing the expectation of the inverse square root by the inverse square root of the concentrated cross-talk energy gives the more detailed approximation
\begin{equation}
C_i(R)\approx
\left[
1+\sum_{j\neq i}K_{ij}^2
+\frac{1}{R}\sum_{j\neq i}\nu_{ij}
\right]^{-1/2}.
\label{eq:detailed-keylaw}
\end{equation}
For the RBF specialization in Eq.~\eqref{eq:rbf-nu}, the simple finite-feature term follows when typical off-target overlaps are weak enough that $K_{ij}^2\ll1$, so that $\nu_{ij}=(1-K_{ij}^2)^2\approx1$ on average. Hence
\begin{equation}
\boxed{
C_i(R)\approx
\left[
1+\frac{N-1}{R}+\sum_{j\neq i}K_{ij}^2
\right]^{-1/2},
}
\end{equation}
which is Eq.~\eqref{eq:keylaw} in the main text. No empirical coefficient is fitted. The approximation is a leading-order consequence of (i) RFF second moments, (ii) weak value cross-correlations, and (iii) concentration of the cross-talk norm. The simple $(N-1)/R$ term is also evaluated empirically under the random-phase normalized RFF used in the broad capacity experiments; the feature-map-specific covariance correction is reserved for the paired construction.

\subsection{Infinite-feature ceiling and reachability}
Taking $R\rightarrow\infty$ in the orthogonal-value model removes the Monte Carlo term but leaves the deterministic kernel overlap:
\begin{equation}
\boxed{
C_{i,\infty}=\left[1+I_{\mathrm{struct},i}\right]^{-1/2},
\qquad
I_{\mathrm{struct},i}=\sum_{j\neq i}K_{ij}^2.
}
\end{equation}
This equality is exact for mutually orthogonal unit values; for finite-dimensional isotropic random values it is the corresponding mean-field limit justified by the concentration statement in Eq.~\eqref{eq:finite-value-concentration}. Therefore, increasing the feature budget cannot overcome the structural ceiling predicted by the key geometry. For a per-item target $C^\star$, the simplified law is reachable only if
\begin{equation}
(C^\star)^{-2}-1-I_{\mathrm{struct},i}>0.
\label{eq:reachability}
\end{equation}
If $R=rN$, the corresponding approximate ratio is
\begin{equation}
r_i^\star
\approx
\frac{1-1/N}{(C^\star)^{-2}-1-I_{\mathrm{struct},i}}.
\label{eq:boundaryclosed}
\end{equation}
When $I_{\mathrm{struct},i}=0$, $N$ is large, and $C^\star=0.95$, Eq.~\eqref{eq:boundaryclosed} gives $r^\star\approx9.26$. In the experiments, we predict the boundary from the \emph{mean} theoretical cosine across heterogeneous items and solve it numerically; Eq.~\eqref{eq:boundaryclosed} is the homogeneous closed-form interpretation.

\section{Joint Key--Value Geometry}\label{app:jointproof}

\subsection{Exact infinite-feature geometry}
For arbitrary unit values, define the exact-kernel retrieval
\begin{equation}
y_{\infty,i}=\sum_j K_{ij}v_j.
\end{equation}
The infinite-feature cosine is therefore exactly
\begin{equation}
C_{i,\infty}
=\frac{\langle y_{\infty,i},v_i\rangle}{\|y_{\infty,i}\|}
=\frac{a_i}{\sqrt{b_i}},
\qquad b_i>0,
\end{equation}
with $a_i=\langle y_{\infty,i},v_i\rangle$ and $b_i=\|y_{\infty,i}\|^2$.
Let $G_{j\ell}=\langle v_j,v_\ell\rangle$. Expanding the two quantities gives
\begin{align}
a_i&=\sum_j K_{ij}G_{ji},\label{eq:a-gram}\\
b_i&=\sum_{j,\ell}K_{ij}K_{i\ell}G_{j\ell}.\label{eq:b-gram}
\end{align}
Equations~\eqref{eq:a-gram}--\eqref{eq:b-gram} are the algebraic reason the aligned/permuted control is informative. Permuting values preserves the multiset of values and hence the marginal value geometry, but it changes how entries of $G$ are paired with the kernel row $K_{i:}$.

\subsection{First-order finite-$R$ approximation}
Let $\delta_{ij}=\hat K_{ij}-K_{ij}$ and define
\begin{equation}
\epsilon_i=\sum_j\delta_{ij}v_j,
\qquad
\hat v_i=y_{\infty,i}+\epsilon_i.
\end{equation}
For unit values, the finite-$R$ cosine is exactly
\begin{equation}
C_i(R)=\frac{a_i+\langle\epsilon_i,v_i\rangle}{\sqrt{b_i+2\langle y_{\infty,i},\epsilon_i\rangle+\|\epsilon_i\|^2}}.
\label{eq:exact-finite-joint}
\end{equation}
For an unbiased feature estimator such as the paired RFF construction, $\mathbb E[\epsilon_i]=0$. Hence
\begin{equation}
\mathbb E\|\hat v_i\|^2=b_i+\mathbb E\|\epsilon_i\|^2,
\end{equation}
while the expected numerator is $a_i$. Replacing the expectation of the ratio in Eq.~\eqref{eq:exact-finite-joint} by the ratio formed from these leading moments is a concentration/mean-field step; in general, $\mathbb E[X/\sqrt{Y}]\neq\mathbb E[X]/\sqrt{\mathbb E[Y]}$. If the finite-feature noise energy is further approximated by $(N-1)/R$, then
\begin{equation}
\boxed{C_i(R)\approx\frac{a_i}{\sqrt{b_i+(N-1)/R}}.}
\end{equation}
This is the joint finite-$R$ law, not an identity for $\mathbb E[C_i(R)]$. For the separately row-normalized random-phase implementation, finite-$R$ unbiasedness is only approximate and the same expression is used as a leading-order prediction. Both $a_i$ and $b_i$ depend on the value Gram matrix, which is why key geometry alone is insufficient for correlated values.

\section{Covariance-Aware Finite-$R$ Noise}\label{app:covproof}

For each independent frequency draw $r$, define
\begin{equation}
c_{ij}^{(r)}=\cos(\omega_r^\top\Delta_{ij}),
\qquad
q_{ir}=\sum_{j\neq i}\big(c_{ij}^{(r)}-K_{ij}\big)v_j.
\end{equation}
Because $\hat K_{ii}=1$, only terms with $j\neq i$ contribute stochastic error, and
\begin{equation}
\epsilon_i=\frac{1}{m}\sum_{r=1}^{m}q_{ir}.
\end{equation}
The $q_{ir}$ are i.i.d. with zero mean. Therefore
\begin{equation}
\mathbb E\|\epsilon_i\|^2
=\frac{1}{m}\mathbb E\|q_{i1}\|^2
=\frac{2D_i}{R},
\qquad
D_i\equiv\mathbb E\|q_{i1}\|^2.
\label{eq:Ddef}
\end{equation}
Let
\begin{equation}
\Gamma_{i,j\ell}
=\operatorname{Cov}_{\omega}
\left(c_{ij},c_{i\ell}\right).
\end{equation}
Expanding Eq.~\eqref{eq:Ddef} gives the exact second-moment identity
\begin{equation}
\boxed{
D_i=\sum_{j,\ell\neq i}\Gamma_{i,j\ell}G_{j\ell}.
}
\label{eq:Dgram}
\end{equation}
For an even shift-invariant kernel, the product-to-sum identity yields
\begin{equation}
\Gamma_{i,j\ell}
=\frac{1}{2}
\left[
\kappa(\Delta_{ij}-\Delta_{i\ell})
+\kappa(\Delta_{ij}+\Delta_{i\ell})
\right]
-K_{ij}K_{i\ell}.
\label{eq:gammaexact}
\end{equation}
Equations~\eqref{eq:Dgram}--\eqref{eq:gammaexact} show that finite-$R$ noise is itself a \emph{joint geometry} quantity: RFF coefficient covariances are weighted by the value Gram matrix. When values are random and approximately orthogonal, $G_{j\ell}\approx\delta_{j\ell}$. For the RBF kernel,
\[
\Gamma_{i,jj}=\frac{1}{2}(1+K_{ij}^{4})-K_{ij}^{2}=\frac{1}{2}(1-K_{ij}^{2})^{2},
\]
so
\[
D_i\approx\frac{1}{2}\sum_{j\neq i}(1-K_{ij}^{2})^{2}\approx\frac{N-1}{2}
\]
only when off-target overlaps are predominantly small. Correlated aligned values can instead make the off-diagonal terms in Eq.~\eqref{eq:Dgram} large.

Replacing the random-value noise reference by Eq.~\eqref{eq:Ddef} gives
\begin{equation}
\boxed{
C_i^{\mathrm{cov}}(R)
\approx
\frac{a_i}{\sqrt{b_i+2D_i/R}}.
}
\end{equation}
This expression is still an expectation-of-ratio approximation: it does not fully retain numerator--denominator coupling or higher moments. This limitation is empirically visible at low $R$ and explains why the covariance denominator improves aligned design-boundary prediction without universally dominating the simpler law for permuted values.

\section{Approximation Status and Scope}\label{app:scope}
The main equations should be interpreted at three levels. First, the identities defining $y_{\infty,i}$, $a_i$, and $b_i$ are exact once the kernel, keys, and values are specified, and $D_i$ is an exact paired-RFF second-moment quantity once the shift-invariant kernel and its spectral construction are specified. Second, the RFF variance and covariance expressions in Appendices~\ref{app:keyproof} and \ref{app:covproof} are exact for the paired sine/cosine feature construction under independent frequency sampling; they are not asserted as exact for the separately row-normalized random-phase feature map used in the broad capacity validation. Third, the key-only law for finite-dimensional random values, the weak-overlap reduction $R^{-1}\sum_{j\neq i}(1-K_{ij}^{2})^{2}\approx(N-1)/R$, and the conversion of finite-$R$ moments into a mean retrieval cosine use concentration/mean-field approximations. The small empirical error in the random-value and high-$R$ regimes is consistent with these approximations in those settings; the correlated-value low-$R$ experiments explicitly document where it becomes less exact. We therefore retain the simpler first-order laws in the main paper and report the higher-order expansion with larger low-$R$ error as a negative result rather than presenting it as a general correction.

\section{Negative Results and Sanity Checks}\label{app:negative}

\subsection{Orthogonal value rotations}
Square orthogonal value codebooks do not create additional representational capacity by themselves. One-hot, signed permutation, Hadamard, and Haar codebooks are related by orthogonal transformations. Apparent differences under AdamW were strongly optimizer/coordinate dependent and largely disappeared under SGD.

\subsection{Learned orthogonal addressing}
Orthogonal learned addressing outperformed Gaussian addressing in some noisy associative-address experiments, but it did not consistently outperform a noise-trained direct MLP. This result motivates studying address geometry but does not support a universal superiority claim.

\subsection{Sparse compositional codes}
Random sparse signed compositional addresses behaved similarly to Gaussian random codes and did not produce a general capacity advantage, so this mechanism is not used in the main analysis.

\subsection{SimilarityCE and soft retrieval}
Similarity cross-entropy improved rotational invariance but weakened alignment for linear Hebbian readout. For square orthogonal address matrices $A$, soft retrieval satisfies
\[
pA(A^\top V)=pV,
\]
so the orthogonal memory cancels algebraically and the operation becomes ordinary value attention.

\subsection{Hybrid objectives}
Combining similarity cross-entropy, cosine alignment, and margin objectives produced a trade-off: better rotational invariance came with weaker linear Hebbian alignment. No tested setting improved both.

\subsection{Higher-order finite-$R$ expansions}
A second-order delta expansion involving additional numerator--denominator covariance terms has larger global error than the first-order approximation, particularly at low $R/N$. It becomes competitive only at large $R$ and is not part of the main theory.

\bibliographystyle{IEEEtran}
\bibliography{references}

\end{document}